\documentclass[10pt,twocolumn,letterpaper]{article}

\usepackage[pagenumbers]{splatwam}

\iflatexml\newenvironment{strip}{}{}\else\usepackage{cuted}\fi
\usepackage{caption}
\usepackage{colortbl}
\usepackage{placeins}
\usepackage{pifont}

\newcommand{\first}{\cellcolor{red!25}}
\newcommand{\second}{\cellcolor{orange!25}}
\newcommand{\third}{\cellcolor{yellow!35}}
\newcommand{\firstnum}[1]{{\setlength{\fboxsep}{0pt}\colorbox{red!25}{\strut\,#1\,}}}
\newcommand{\secondnum}[1]{{\setlength{\fboxsep}{0pt}\colorbox{orange!25}{\strut\,#1\,}}}
\newcommand{\thirdnum}[1]{{\setlength{\fboxsep}{0pt}\colorbox{yellow!35}{\strut\,#1\,}}}
\newcommand{\cmark}{\checkmark}
\newcommand{\xmark}{$\times$}

\definecolor{linkblue}{rgb}{0.21,0.49,0.74}
\usepackage[pagebackref,breaklinks,colorlinks,allcolors=linkblue]{hyperref}
\hypersetup{pdftitle={SplatWAM: Splatting World Action Models for Real-Time 3D Imagination},pdfauthor={Frank Tsu-Fang Lu, Karthik Sundar Senthilkumar, Christopher Thirgood, Oscar Mendez, Simon Hadfield}}

\title{SplatWAM: Splatting World Action Models for Real-Time 3D Imagination}

\author{
Frank Tsu-Fang Lu$^{1}$ \quad Karthik Sundar Senthilkumar$^{1}$ \quad Christopher Thirgood$^{1}$\\
Oscar Mendez$^{2}$ \quad Simon Hadfield$^{1}$\\[0.5em]
\iflatexml
$^{1}$CVSSP, University of Surrey \quad $^{2}$Locus Robotics\\
{\tt\footnotesize \{tsufanglu,k.senthilkumar,c.thirgood,s.hadfield\}@surrey.ac.uk}\\
{\tt\footnotesize omendez@locusrobotics.com}
\else
\begin{tabular}{@{}c@{\hspace{2.5em}}c@{}}
$^{1}$CVSSP, University of Surrey & $^{2}$Locus Robotics\\
{\tt\footnotesize \{tsufanglu,k.senthilkumar,c.thirgood,s.hadfield\}@surrey.ac.uk} & {\tt\footnotesize omendez@locusrobotics.com}
\end{tabular}\fi\\[0.3em]
{\footnotesize\url{https://frankcholula.com/splatwam}}
}

\begin{document}
\maketitle
\begin{strip}
\centering
\includegraphics[width=\textwidth]{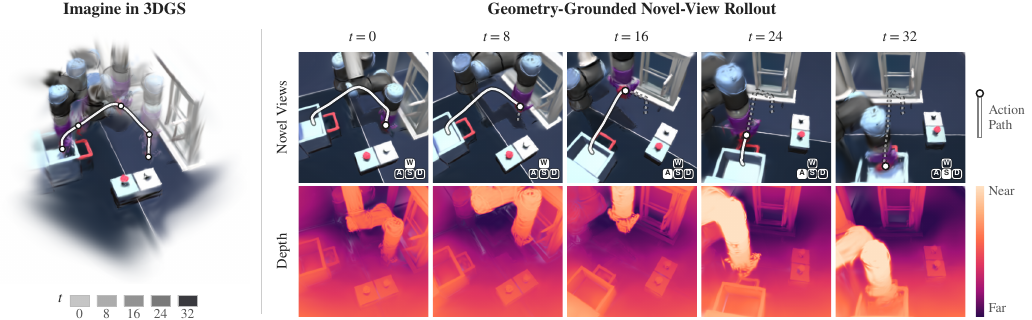}
\captionsetup{hypcap=false}
\captionof{figure}{\textbf{SplatWAM imagines in 3D.} SplatWAM predicts future 3D Gaussian states, rasterizes them from any viewpoint, and outputs actions from the same states. We show a 32-step rollout with a sweeping camera, with rendered depth showing consistent geometry.}
\captionsetup{hypcap=true}
\label{fig:teaser}
\end{strip}

\begin{abstract}
World models typically predict future observations in pixel or latent space, limiting their ability to directly represent true 3D scene evolution.
We present \textbf{SplatWAM} (\textbf{Splat}ting \textbf{W}orld \textbf{A}ction \textbf{M}odel), a generative world model that learns action-conditioned dynamics directly over compact 3D Gaussian primitives. A tokenizer SplatWAM-$\mathcal{T}$ first encodes a variable number of unposed views into a fixed budget of Gaussian tokens. A dynamics model SplatWAM-$\mathcal{D}$ then predicts future Gaussian states in a single forward pass per action, using log-Euclidean covariance coordinates and differentiable rendering supervision to facilitate learning in an explicit 3D space. Each predicted state can be rasterized from arbitrary viewpoints, enforcing a shared 3D representation across views by construction. On OGBench Scene, SplatWAM renders novel views $12\times$ faster than video world models (${\sim}$80 FPS on an RTX 5090) with up to 2.0 dB higher PSNR.
We further introduce SplatWAM-$\mathcal{P}$, a lightweight flow-matching policy head jointly trained with the dynamics model. Across 8 task suites, model-free Gaussian-token policies outperform pixel-based counterparts by up to 24.8 percentage points. SplatWAM-$\mathcal{P}$ improves over Latent Particle World Model (LPWM) by 12 percentage points and matches HIQL without a learned value function. On LIBERO, our ${\sim}$1B model outperforms WorldVLA (${\sim}$7B) and the robot-pretrained OpenVLA (${\sim}$7B), while matching OpenVLA-OFT when both are trained without large-scale robot pretraining. Code and checkpoints will be released.
\end{abstract}

\section{Introduction}
\label{sec:intro}

Video generation models have become the dominant route to world modeling~\cite{genie,unisim,cosmos}.
However, these models are pixel-first and may produce geometrically impossible predictions (\S\ref{sec:exp_sim}, Fig.~\ref{fig:counterfactual}).
To obtain an explorable world state for downstream planning or control, a second stage of feed-forward splatting~\cite{pixelsplat,splatt3r} or per-scene lifting~\cite{3dgs} is required.
Even recent 3D-aware variants inject geometry only as conditioning, \eg, a rendered point-cloud cache~\cite{gen3c} or a latent spatial memory~\cite{mirage}.
The output remains video, and every new viewpoint requires another sampling pass, often with multi-billion-parameter models trained on thousands of GPUs~\cite{cosmos}.
This view dependence also affects control, as pixel-based policies degrade under camera shifts~\cite{knowcamera,mvdemos}.
Multi-view supervision~\cite{mvmwm} and geometric pretraining~\cite{geo3drep} reduce this sensitivity, but use the extra views and geometry only to train the representation, not as the state the model predicts. Because the state is never geometrically grounded, consistency has to be learned rather than enforced, even when several views are predicted jointly~\cite{ctrlworld}.

Without a shared 3D state, independently generated views may disagree in geometry, occlusion, and parallax, with inconsistencies compounding over long rollouts. Camera-controllable video diffusion~\cite{cameractrl,gen3c} takes viewpoint as input, but fixes the camera trajectory before generation. Latent world models~\cite{dreamerv3,tdmpc2,vjepa2,lewm} avoid pixel generation but do not directly expose a 3D state, requiring additional decoding to support novel-view rendering, and therefore do not explicitly enforce geometric consistency.

At the other end of the spectrum, the 4D Gaussian splatting lineage~\cite{4dgs,yang4dgs,deformable3dgs} keeps an explicit rasterizable state and models motion directly, but it is a per-scene optimization that densifies to many primitives.
With no generative prior, these methods primarily reconstruct observed motion rather than predict controllable future scene states.
Feed-forward splatting~\cite{pixelsplat,noposplat} avoids per-scene optimization but yields a static snapshot.

These existing approaches either learn dynamics in compressed latent spaces or rely on per-scene reconstruction. We instead propose SplatWAM, an action-conditioned world model that directly predicts compact 3D Gaussian states from a feed-forward representation of unposed RGB views. The resulting states can be rasterized from arbitrary viewpoints without a learned latent decoder (Fig.~\ref{fig:teaser}).
We evaluate on two robot manipulation benchmarks: OGBench~\cite{ogbench} for goal-conditioned and LIBERO~\cite{libero} for language-conditioned behavior cloning. Our contributions are as follows:
\begin{itemize}
\item \textbf{A compact, geometry-grounded 3D state.}
SplatWAM-$\mathcal{T}$ maps a variable number of unposed context views to a fixed budget of Gaussian tokens. As policy inputs, these tokens outperform pixels by up to 24.8 percentage points.
\item \textbf{Direct generative dynamics in Gaussian space.}
SplatWAM-$\mathcal{D}$ learns action-conditioned transitions directly over compact Gaussian states, without a separate learned latent compression or decoding stage. Gaussian-specific covariance parameterization and differentiable rendering supervision support prediction in this 3D space.
It rolls out at $\sim$80 FPS on a single RTX 5090, and any predicted state can be rendered from any viewpoint.
\item \textbf{Joint prediction and control with a shared world-model backbone.} SplatWAM-$\mathcal{P}$ adds agent tokens and a flow-matching action head to SplatWAM-$\mathcal{D}$, jointly optimizing prediction and control. It outperforms LPWM~\cite{lpwm}, matches HIQL~\cite{hiql} on OGBench Scene, and is competitive with pretrained vision-language-action models (VLAs) on LIBERO without robot pretraining.
\end{itemize}

\section{Related work}
\label{sec:related}

\subsection{World models}
Pixel-space world models predict future frames autoregressively over VQ tokens~\cite{ivideogpt} or by adapting a video diffusion prior~\cite{ctrlworld}. Latent world models such as LPWM~\cite{lpwm} model stochastic particle dynamics over discovered object keypoints, which supports goal-conditioned control but does not encode geometry and cannot be rendered from new viewpoints.
Driving world models~\cite{occworld,copilot4d} predict 3D occupancy or LiDAR scenes but are not photorealistically renderable. Gaussian World Model (GWM)~\cite{gwm} reconstructs per-pixel Gaussians feed-forward, but their large number of primitives forces it to downsample and compress them into a latent in which the dynamics are learned. As a result, its predictions degrade sharply once rasterized (Fig.~\ref{fig:rollout}b). PointWorld~\cite{pointworld} scales action-conditioned 3D prediction to in-the-wild scenes but regresses deterministic per-point flow, and MRO-GWM~\cite{mrogwm} likewise regresses rigid per-object SE(3) motion, relying on object masks and posed RGB-D. Prior world models thus cannot render their state from new viewpoints, must decode it first, or regress motion deterministically. SplatWAM instead predicts Gaussian parameters directly and generatively, so its state is rendered without a decoder, can be inspected from any viewpoint, and is supervised by dense differentiable rendering without the need for masks and poses.

\begin{figure*}[t]
\centering
\includegraphics[width=\textwidth]{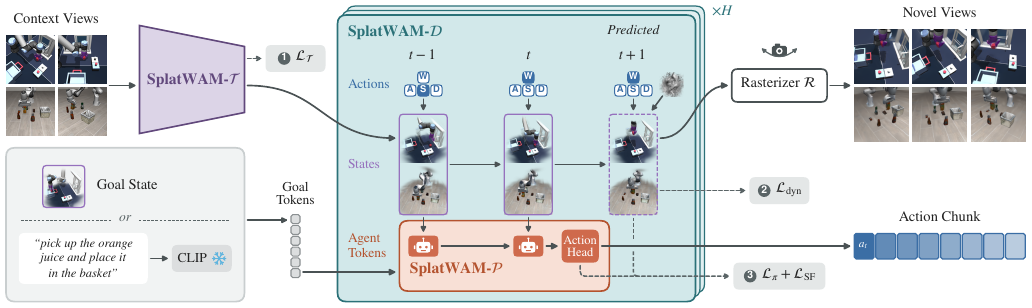}
\caption{\textbf{SplatWAM overview.} SplatWAM-$\mathcal{T}$ lifts unposed context views into Gaussian states. SplatWAM-$\mathcal{D}$ denoises the next state given the action, repeated $H$ times. SplatWAM-$\mathcal{P}$'s agent tokens read the observed states and goal tokens to output an action chunk. Dashed arrows are losses: \ding{182} trains $\mathcal{T}$; \ding{183} trains $\mathcal{D}$ with $\mathcal{T}$ frozen; \ding{184} then trains $\mathcal{D}$ and $\mathcal{P}$ jointly with $\mathcal{L}_{\pi}+\mathcal{L}_{\mathrm{SF}}$.}
\label{fig:arch}
\end{figure*}

\subsection{World action models}
World action models~\cite{dreamzero} predict future states and actions in a single model.
Unified World Models~\cite{uwm} couple video and action diffusion, and WorldVLA~\cite{worldvla} unifies a VLA and an image world model in one autoregressive model. Video Prediction Policy~\cite{vpp} conditions an inverse-dynamics policy on the future representations of a fine-tuned video diffusion model. Among 3D-aware models, ManiGaussian~\cite{manigaussian,manigaussianpp} and GAF~\cite{gaf} predict per-Gaussian motion for a single step, as auxiliary supervision for the policy or to infer actions.
GaussianDream~\cite{gaussiandream} plugs into a pretrained VLA whose future head is conditioned on a horizon embedding rather than on actions and discards its decoders at deployment.
4DGS-WAM~\cite{4dgswam} maintains a renderable 4D state whose dynamics factorize into independently transformed objects over a static background.
The closest approach to SplatWAM is the structured 4D latent predictive model~\cite{structured4d}, which flow-matches a sparse voxel latent, then decodes it to Gaussians for rendering and to points for control, but requires posed multiview RGB-D. Unlike these methods, SplatWAM predicts multi-step generative dynamics directly over Gaussian states inferred from unposed RGB, which also serve as policy inputs.

\section{Methodology}
\label{sec:method}

Given a set of unposed RGB views $\mathcal{I}_t = \{I_t^v\}_{v \in \mathcal{C}}$ of a dynamic scene at time $t$, where $\mathcal{C}$ indexes the context views, SplatWAM represents the scene as $N$ explicit Gaussian primitives,
\begin{equation}
s_t = (x_{t,1},\dots,x_{t,N}), \qquad x_{t,i} = (\mu_i, q_i, \varsigma_i, \alpha_i, c_i) \in \mathbb{R}^{14},
\end{equation}
with position $\mu_i \in \mathbb{R}^3$, rotation quaternion $q_i \in \mathbb{R}^4$, scale $\varsigma_i \in \mathbb{R}^3$, opacity $\alpha_i \in \mathbb{R}$, and color $c_i \in \mathbb{R}^3$, mapped to a valid Gaussian before rendering (App.~\ref{sec:app_tok_impl}).
For any camera $\pi^v$, a differentiable rasterizer $\mathcal{R}$ renders the image $\hat{I}^v = \mathcal{R}(s_t, \pi^v)$.

SplatWAM has three learned components (Fig.~\ref{fig:arch}).
A feed-forward tokenizer, trained with a novel-view-synthesis objective, lifts the context views into an explicit Gaussian scene state,
\begin{equation}
\operatorname{SplatWAM}\text{-}\mathcal{T}:\; \mathcal{I}_{t}\longmapsto s_t.
\end{equation}
Each $\mathcal{I}_t$ is a partial observation of the scene, with $\mathcal{R}$ as the observation model. A single estimate $s_t$ provides only a snapshot of the scene, offering no explicit temporal context for inferring its dynamics. Hence, SplatWAM-$\mathcal{D}$ and SplatWAM-$\mathcal{P}$ condition on a history of past states.
SplatWAM-$\mathcal{D}$ then predicts future states from this history and the actions, following the controllable world model formulation of Hou \etal~\cite{wmsurvey},
\begin{equation}
\begin{split}
&\operatorname{SplatWAM}\text{-}\mathcal{D}:\; \bigl(s_{\le t}, a_{<t+H}\bigr) \longmapsto s_{t+1:t+H} \\
&\qquad \sim p_\theta\!\left(s_{t+1:t+H} \mid s_{\le t}, a_{<t+H}\right),
\end{split}
\end{equation}
where $H$ is the rollout horizon, $a_t$ transitions $s_t$ to $s_{t+1}$, and $a_{t:t+H-1}$ produces the Gaussian-state trajectory $s_{t+1:t+H}$.
SplatWAM-$\mathcal{D}$ generates this trajectory autoregressively, sampling $s_{t+1}\sim p_\theta(s_{t+1}\mid s_{\le t},a_{\le t})$ and appending it to the history before the next step.
Since the rasterizer is deterministic, all stochasticity resides in $\mathcal{D}$.
Finally, a goal-conditioned policy acts on the Gaussian state directly,
\begin{equation}
\operatorname{SplatWAM}\text{-}\mathcal{P}: (s_{\leq{t}}, g) \longmapsto a_{t:t+k-1},
\end{equation}
where the goal $g$ is either a goal state ($s_g = \operatorname{SplatWAM}\text{-}\mathcal{T}(\mathcal{I}_g)$ tokenized from goal images) or a language instruction, and the policy outputs a chunk of $k$ actions.
Latent 3D world models~\cite{gwm,structured4d} decode to Gaussians only for rendering. SplatWAM has no decoder: $\mathcal{T}$ encodes, $\mathcal{D}$ predicts, $\mathcal{R}$ renders, and $\mathcal{P}$ reads the same $N$ Gaussians, enabling direct image-space supervision (\S\ref{sec:gaussiandyn}).

\subsection{Explicit 3D representation with SplatWAM-\texorpdfstring{$\mathcal{T}$}{T}}
\label{sec:tokenizer}

SplatWAM-$\mathcal{T}$ follows the fixed-budget design of C3G~\cite{c3g}.
A VGGT backbone~\cite{vggt}, fine-tuned on our data, encodes the unposed context views into patch tokens, to which we append a set of $N$ learned query tokens $Q = (Q_1, \dots, Q_N)$.
A small transformer applies self-attention over this joint sequence and a linear head $h$ decodes the output at each query position into one Gaussian primitive, expressed in a canonical frame anchored to the first context view:
\begin{equation}
\begin{split}
s_t &= \operatorname{SplatWAM}\text{-}\mathcal{T}(\mathcal{I}_t) \\
&= h\big(\operatorname{Attn}(\operatorname{VGGT}(\mathcal{I}_t),\, Q)\big),
\end{split}
\end{equation}
where $\operatorname{Attn}$ returns the transformed query tokens.
The queries are shared across scenes and time steps, so primitive $x_{t,i}$ is always decoded from query $Q_i$. This establishes a fixed indexing convention across frames without requiring explicit physical point correspondence. This fixed query budget decouples token count from image resolution and number of views, keeping the state size constant and making downstream dynamics learning tractable.

We train SplatWAM-$\mathcal{T}$ once for all downstream tasks using a novel-view synthesis objective~\cite{pixelsplat,c3g}. Given context views $\mathcal{C}$ and a held-out view $v^*$, we render $\mathcal{V}=\mathcal{C}\cup\{v^*\}$ and minimize
\begin{equation}
\mathcal{L}_{\mathcal{T}} = \mathcal{L}_{\mathrm{photo}} + \mathcal{L}_{\mathrm{geom}} + \mathcal{L}_{\mathrm{reg}},
\end{equation}
where the photometric loss $\mathcal{L}_{\mathrm{photo}}$ compares rendered and ground-truth images, $\mathcal{L}_{\mathrm{geom}}$ supervises rendered depth and normals, and $\mathcal{L}_{\mathrm{reg}}$ keeps the primitives well-formed.
All image-space terms are averaged over the views in $\mathcal{V}$.
We sample the number of context views at random during training.
Training details and qualitative reconstructions are in Apps.~\ref{sec:app_tok_impl} and~\ref{sec:app_repr}.

Let $I^v$ be the ground-truth image at view $v$ and $\hat{I}^v=\mathcal{R}(s_t,\pi^v)$ the equivalent rendering of the 3DGS model. We combine a pixel-wise reconstruction loss with a perceptual loss,
\begin{equation}
\mathcal{L}_{\mathrm{photo}} = \mathbb{E}_{v,p}\big[\|\hat{I}^v_p-I^v_p\|_2^2\big] + \lambda_p\,\mathbb{E}_v\big[\operatorname{LPIPS}(\hat{I}^v,I^v)\big],
\end{equation}
where the first term averages over all pixels and views in $\mathcal{V}$, and $\lambda_p$ weights the perceptual term.

Photometric losses can be minimized despite incorrect geometry, but reliable rollouts require geometrically grounded scene states, particularly for interactions involving contact and occlusion (\S\ref{sec:exp_sim}). We therefore distill depth and normals from the pretrained backbone's depth prediction $\bar{D}^v$.
Let $\hat{D}^v$ be the depth rasterized from $s_t$ at view $v$, rescaled by the median ratio $\rho_v=\operatorname{med}(\bar{D}^v)/\operatorname{med}(\hat{D}^v)$ to absorb the teacher's unknown scale.
We then minimize
\begin{equation}
\mathcal{L}_{\mathrm{geom}} = \mathbb{E}_{v}\,\mathbb{E}_{p}\big[\lambda_d\, |\rho_v \hat{D}^v_p - \bar{D}^v_p| + \lambda_n\, (1 - \hat{\mathbf{n}}^v_p \cdot \bar{\mathbf{n}}^v_p)\big],
\end{equation}
where $\lambda_d$ and $\lambda_n$ weight the two terms.
Both normal maps are computed from their depth maps by back-projecting each pixel and taking the normalized cross product of the pointmap's Sobel gradients.
SplatWAM-$\mathcal{T}$, trained with this loss, has the lowest depth and normal error in Table~\ref{tab:ffgs}.

Compact feed-forward models trained from sparse views can drift into degenerate primitives, such as translucent needles that contribute negligible mass. We therefore regularize opacity coverage, anisotropy, and covariance effective rank~\cite{erankgs,physgaussian}:
\begin{equation}
\mathcal{L}_{\mathrm{reg}} = \lambda_\alpha \mathcal{L}_\alpha + \lambda_{\mathrm{aniso}} \mathcal{L}_{\mathrm{aniso}} + \lambda_{\mathrm{erank}} \mathcal{L}_{\mathrm{erank}}.
\end{equation}
The coverage term encourages the primitives to cover the full image,
\begin{equation}
\mathcal{L}_\alpha = \mathbb{E}_{v,p}\big[1 - A_p^v\big],
\end{equation}
where $A_p^v \in [0,1]$ is the accumulated alpha at pixel $p$ in rendered view $v$.
The anisotropy term limits extreme elongation, and the effective-rank term penalizes near-rank-one primitives (App.~\ref{sec:app_tok_impl}).
Together, they suppress needle-like degeneracies with little effect on photometric quality.

\subsection{World modeling with SplatWAM-\texorpdfstring{$\mathcal{D}$}{D}}
\label{sec:dynamics}

SplatWAM-$\mathcal{D}$ predicts future Gaussian states directly.
It operates on $\tilde{s}_t$, the same state with each covariance in log-Euclidean coordinates (\S\ref{sec:gaussiandyn}).
For a training window $\tilde{s}_{1:T}$, we implement diffusion forcing~\cite{diffusionforcing}, independently sampling a signal level $\tau_t$ for each timestep. Each state is interpolated with Gaussian noise according to its assigned $\tau_t$:
\begin{equation}
z_t=\tau_t \tilde{s}_t+(1-\tau_t)\epsilon_t,\quad
\epsilon_t\sim\mathcal{N}(0,I).
\label{eq:df}
\end{equation}
A causal transformer $f_\theta$ then predicts the clean states:
\begin{equation}
\hat{\tilde{s}}_{1:T}=f_\theta(z_{1:T},\tau_{1:T},d_{1:T},a_{0:T-1}).
\end{equation}

Noising each frame independently exposes the model to imperfect context during training, which alleviates training-inference mismatch~\cite{scheduledsampling,exposurebias} when conditioning on its own predictions during autoregressive rollouts.
Following Li and He~\cite{jit}, we use $x$-prediction to mitigate error accumulation over long rollouts while producing Gaussian states that can be rendered directly for supervision.
We adopt the shortcut-forcing objective $\mathcal{L}_{\mathrm{SF}}$ from Dreamer 4~\cite{dreamer4}, which combines a flow-matching loss with a shortcut-consistency loss that trains one large denoising step to match two smaller steps~\cite{shortcut}.
We apply classifier-free-guidance action dropout~\cite{cfg}, replacing all actions in a random subset of training sequences with a learned null action token. As a result, SplatWAM-$\mathcal{D}$ also learns an action-marginalized future. SplatWAM-$\mathcal{P}$ reuses this token (\S\ref{sec:policy}).
The full objective and training details are in Apps.~\ref{sec:app_sf} and~\ref{sec:app_wm_impl}. Each frame is represented by $N$ Gaussian tokens, together with tokens for the preceding action, $\tau_t$, step size $d_t$, and register tokens~\cite{registers}.
The transformer applies within-frame self-attention at every layer and causal temporal attention across frames for each primitive at interleaved layers.

At inference, each future frame is initialized from noise and denoised in $K=1/d_t$ shortcut steps.
Observed context frames are provided clean and assigned the highest signal level $\tau=1-d_{\min}$, whereas generated frames are re-noised to the context signal level $\tau_{\mathrm{ctx}}$ before conditioning subsequent predictions.
For rollouts beyond the training window, the model retains at most the most recent $T-1$ states, imposing a $(T{-}1)$th-order Markov approximation, and re-indexes their temporal positions.

\subsubsection{Gaussian-specific dynamics modeling}
\label{sec:gaussiandyn}

Explicit Gaussians differ from patch tokens in their continuous geometric parameters and fixed token order. Modeling their dynamics hence requires consideration of spatial variation, parameter ambiguity, and the temporal evolution of Gaussian states. We introduce three modifications to make these dynamics easier to learn: a 3D positional embedding, a rendering loss, and a covariance reparameterization.

Each primitive keeps the same token index in every frame.
Unlike a patch index, this index does not correspond to a fixed spatial location.
We therefore assign each index a fixed sinusoidal embedding computed from its mean position over the training set, $\bar{\mu}_i$, making tokens addressable by where they typically lie in 3D space.
This mean is informative because primitive $i$ is always decoded from the same learned query $Q_i$ in the canonical frame of the first context view (\S\ref{sec:tokenizer}).

Because $s_t$ can be differentiably rasterized, we complement the shortcut-forcing objective with an image-space consistency loss.
Specifically, we rasterize the predicted final frame $\hat{\tilde{s}}_T$ at one held-out viewpoint $\pi^v$ and compare it with a rendering of the target state $\tilde{s}_T$, tokenized from the observed frame, from the same viewpoint,
\begin{equation}
\mathcal{L}_{\mathrm{render}}=\big\|\mathcal{R}(\hat{\tilde{s}}_T,\pi^v)-\mathcal{R}(\tilde{s}_T,\pi^v)\big\|_2^2,
\end{equation}
and minimize
\begin{equation}
\mathcal{L}_{\mathrm{dyn}}=\mathcal{L}_{\mathrm{SF}}+\lambda_{\mathrm{render}}\,\mathcal{L}_{\mathrm{render}},
\label{eq:dyn}
\end{equation}
where $\lambda_{\mathrm{render}}$ weights the rendering term.
This supervises the rendered appearance of predicted states, providing complementary constraints on Gaussian geometry, scale, and opacity.

SplatWAM-$\mathcal{T}$ outputs a rotation quaternion $q_i$ and scale $\varsigma_i$ for each Gaussian.
Together, they define its covariance
\begin{equation}
\Sigma_i=R(q_i)\operatorname{diag}(\varsigma_i)^2R(q_i)^\top.
\end{equation}
The quaternion--scale factorization is non-unique: $q_i$ and $-q_i$ encode the same rotation, while orientation becomes ambiguous for nearly isotropic Gaussians.
We instead represent covariance in six-dimensional log-Euclidean coordinates~\cite{logeuclidean}, centered at the training log-Euclidean Fr\'echet mean $\bar{\Sigma}$:
\begin{equation}
\nu_i=\operatorname{vech}\!\left(\log\Sigma_i-\log\bar{\Sigma}\right)\in\mathbb{R}^6,
\end{equation}
where $\operatorname{vech}$ extracts the six unique entries of a symmetric matrix, scaling off-diagonal terms by $\sqrt{2}$.
Each primitive is represented as $\tilde{x}_{t,i}=(\mu_i,\nu_i,\alpha_i,c_i)\in\mathbb{R}^{13}$, forming the state $\tilde{s}_t=(\tilde{x}_{t,1},\dots,\tilde{x}_{t,N})$ of Eq.~\eqref{eq:df}.
For rendering, we recover
\begin{equation}
\Sigma_i=\exp\!\left(\log\bar{\Sigma}+\operatorname{vech}^{-1}(\nu_i)\right),
\end{equation}
which is symmetric positive definite by construction.

\subsection{Behavior cloning with SplatWAM-\texorpdfstring{$\mathcal{P}$}{P}}
\label{sec:policy}
We attach a lightweight policy head to SplatWAM-$\mathcal{D}$.
The policy runs its own pass of SplatWAM-$\mathcal{D}$ over clean observed states, with a small set of learned agent tokens $\pi$ appended to each frame.
At each layer, the agent tokens cross-attend to the current frame's world tokens, attend causally over their own history, and read goal tokens $G$: the Gaussian tokens of a goal state $s_g$ on OGBench, or the per-token outputs of a frozen CLIP~\cite{clip} text encoder for the instruction on LIBERO.
An asymmetric within-frame attention mask (App.~\ref{sec:app_pol_impl}) prevents world tokens from attending to $\pi$ or $G$, keeping the dynamics task-agnostic.
We observe copycat behavior when exposing past actions to the policy~\cite{causalconfusion,copycat}, motivating our use of the learned null action token (\S\ref{sec:dynamics}) in the policy pass, while the dynamics pass retains action conditioning.

SplatWAM-$\mathcal{P}$ generates the $k$-step action chunk $\mathbf{a}_t=a_{t:t+k-1}$ of \S\ref{sec:method} using a flow-matching head $u_\theta$ conditioned on the agent tokens at frame $t$.
We sample a flow time $\eta\sim\mathcal{U}[0,1]$ and noise $\epsilon\sim\mathcal{N}(0,I)$, form $\mathbf{a}^{\eta}_t=\eta\,\mathbf{a}_t+(1-\eta)\epsilon$, and train the head to predict the velocity toward the clean chunk,
\begin{equation}
\mathcal{L}_{\pi}=\mathbb{E}\Bigl[\bigl\|u_\theta(\mathbf{a}^{\eta}_t,\eta\mid s_{\le t},G)-(\mathbf{a}_t-\epsilon)\bigr\|_2^2\Bigr],
\end{equation}
where $\theta$ denotes all jointly trained parameters, including the agent tokens and the head.
We first train SplatWAM-$\mathcal{D}$ alone, then jointly optimize it with SplatWAM-$\mathcal{P}$ using
\begin{equation}
\mathcal{L}_{\mathrm{joint}}=\mathcal{L}_{\pi}+\mathcal{L}_{\mathrm{SF}},
\end{equation}
where $\mathcal{L}_{\mathrm{SF}}$ is the shortcut-forcing objective (App.~\ref{sec:app_sf}), computed on the same windows with fresh noise. We ablate keeping the rendering loss ($\mathcal{L}_{\pi}+\mathcal{L}_{\mathrm{dyn}}$) and insulating action-head gradients ($\mathcal{L}^{\mathrm{sg}}_{\pi}+\mathcal{L}_{\mathrm{dyn}}$)~\cite{knowinsulate} in App.~\ref{sec:app_joint}. On OGBench, goals are relabeled in hindsight from play data~\cite{her,gcsl}. Training and inference details are in App.~\ref{sec:app_pol_impl}.

\section{Experiments}
\label{sec:experiments}
\begin{table*}[t]
\centering
\small
\caption{\textbf{Novel-view appearance and geometry on OGBench Scene using two context views.} Colors mark the \firstnum{best}, \secondnum{second-best}, and \thirdnum{third-best} results. \#G is the primitive count. AbsRel and $\delta_1$ measure depth error and accuracy, Normal the surface-normal angular error.}
\label{tab:ffgs}
\begin{tabular}{clccccccc}
\toprule
Pose-Free & Method & \#G & PSNR$\uparrow$ & SSIM$\uparrow$ & LPIPS$\downarrow$ & AbsRel$\downarrow$ & $\delta_1\uparrow$ & Normal\,($^\circ$)$\downarrow$ \\
\midrule
\xmark & pixelSplat~\cite{pixelsplat} & \second 100K & \first 24.52 & \first .884 & \first .097 & .145 & 81.2 & 48.9 \\
\xmark & MVSplat~\cite{mvsplat} & \second 100K & \second 23.46 & \second .874 & \first .097 & \third .095 & \third 89.7 & \third 46.4 \\
\cmark & AnySplat~\cite{anysplat} & \third $\sim$370K & 22.73 & \third .848 & \first .097 & \second .059 & \first 96.8 & \second 42.5 \\
\cmark & C3G~\cite{c3g} & \first 2K & 21.04 & .789 & \third .166 & .131 & 83.1 & 52.6 \\
\cmark & \textbf{SplatWAM-$\mathcal{T}$ (Ours)} & \first 2K & \third 23.00 & .837 & \second .149 & \first .055 & \second 95.0 & \first 37.9 \\
\bottomrule
\end{tabular}
\end{table*}
\begin{table*}[t]
\centering
\iflatexml\scriptsize\setlength{\tabcolsep}{1.5pt}\else\footnotesize\setlength{\tabcolsep}{0pt}\fi
\caption{\textbf{Action-conditioned rollout quality on OGBench Scene.} Cells report results at horizons $H=16$/$H=32$. Colors mark the \firstnum{best}, \secondnum{second-best}, and \thirdnum{third-best} results. $^{\dagger}$Novel views are lifted post-hoc with SplatWAM-$\mathcal{T}$. GWM and SplatWAM-$\mathcal{D}$ are rasterized.}
\label{tab:rollout}
\begin{tabular*}{\textwidth}{@{\extracolsep{\fill}}lccccccccccc@{}}
\toprule
& \multicolumn{5}{c}{\textbf{Input Views}} & \multicolumn{6}{c}{\textbf{Novel Views}} \\
\cmidrule(r){2-6}\cmidrule(l){7-12}
Method & PSNR$\uparrow$ & SSIM$\uparrow$ & LPIPS$\downarrow$ & FVD$\downarrow$ & FPS$\uparrow$ & PSNR$\uparrow$ & SSIM$\uparrow$ & LPIPS$\downarrow$ & FVD$\downarrow$ & MEt3R$\downarrow$ & FPS$\uparrow$ \\
\midrule
iVideoGPT$^{\dagger}$~\cite{ivideogpt} & \thirdnum{20.56}/\thirdnum{19.28} & \thirdnum{.838}/\thirdnum{.812} & \thirdnum{.093}/\thirdnum{.120} & \secondnum{49}/\secondnum{75} & 11.0/11.0 & \thirdnum{19.13}/\thirdnum{18.19} & \thirdnum{.760}/\thirdnum{.734} & \thirdnum{.157}/\thirdnum{.184} & \thirdnum{257}/\thirdnum{506} & \secondnum{.124}/\secondnum{.126} & 4.4/4.4 \\
Cosmos3-Edge$^{\dagger}$~\cite{cosmos3} & \firstnum{25.30}/\secondnum{23.73} & \firstnum{.908}/\firstnum{.884} & \firstnum{.046}/\firstnum{.061} & \firstnum{26}/\firstnum{39} & \thirdnum{18.9}/\thirdnum{18.9} & \secondnum{21.82}/\secondnum{21.03} & \secondnum{.831}/\secondnum{.808} & \secondnum{.113}/\secondnum{.126} & \secondnum{194}/\secondnum{355} & \firstnum{.122}/\firstnum{.123} & \thirdnum{6.6}/\thirdnum{6.6} \\
GWM~\cite{gwm} & 19.57/17.91 & .784/.733 & .205/.256 & 397/733 & \secondnum{54.6}/\secondnum{56.5} & 9.18/9.10 & .222/.212 & .654/.661 & 4118/5085 & \thirdnum{.198}/\thirdnum{.195} & \secondnum{43.3}/\secondnum{44.4} \\
\textbf{SplatWAM-$\mathcal{D}$ (Ours)} & \secondnum{24.57}/\firstnum{24.15} & \secondnum{.883}/\secondnum{.874} & \secondnum{.085}/\secondnum{.088} & \thirdnum{173}/\thirdnum{338} & \firstnum{83.8}/\firstnum{83.9} & \firstnum{23.37}/\firstnum{23.02} & \firstnum{.871}/\firstnum{.863} & \firstnum{.099}/\firstnum{.102} & \firstnum{181}/\firstnum{313} & \firstnum{.122}/\firstnum{.123} & \firstnum{78.8}/\firstnum{78.8} \\
\bottomrule
\end{tabular*}
\end{table*}

Our experiments evaluate SplatWAM at three levels, representation, dynamics, and control, followed by ablations.
They are organized around four questions.
\begin{itemize}
\item \textbf{Representation (\S\ref{sec:exp_repr}).} Does a compact Gaussian state preserve scene appearance and geometry?
\item \textbf{Dynamics (\S\ref{sec:exp_sim}).} Does an explicit 3D state make action-conditioned rollouts more view-consistent and physically faithful than pixel- and latent-space world models?
\item \textbf{Control (\S\ref{sec:exp_control}).} How do Gaussian representations compare with pixel-based observations for control, and can a pretrained 3D world-model backbone be effectively reused for action prediction?
\item \textbf{Ablations (\S\ref{sec:exp_ablation}).} How do the 3D positional embedding, rendering supervision, and covariance reparameterization affect rollout quality?
\end{itemize}

We evaluate SplatWAM on OGBench~\cite{ogbench} and LIBERO~\cite{libero}. For OGBench, we train on the official Scene and Cube play datasets and evaluate control in their corresponding visual environments across five official tasks. For LIBERO, we pretrain the tokenizer on Spatial, Object, Goal, and 90, holding out Long to evaluate out-of-distribution representation transfer. World models and policies are then trained on each suite's demonstrations following OpenVLA's evaluation protocol. Training, evaluation splits, and baselines for each experiment are specified in the corresponding subsection.

For reconstruction and rollout experiments, we replay simulator states and render nine cameras along a fixed arc around the scene (App.~\ref{sec:app_tok_eval}). Five are context cameras that include each benchmark's native viewpoints, and the other four lie between them and serve only as held-out rendering targets. All comparisons use only two context views.

\subsection{SplatWAM-\texorpdfstring{$\mathcal{T}$}{T}'s explicit representation}
\label{sec:exp_repr}
We compare against pixel-aligned feed-forward splatting from posed views (pixelSplat~\cite{pixelsplat}, MVSplat~\cite{mvsplat}) and from unposed views (AnySplat~\cite{anysplat}), and against C3G~\cite{c3g}, which also decodes a fixed budget of learned queries into Gaussians.
Evaluation details are in App.~\ref{sec:app_tok_eval}. SplatWAM-$\mathcal{T}$'s advantage is its geometry and compactness rather than rendering quality (Table~\ref{tab:ffgs}, Fig.~\ref{fig:app_qual}). With the same primitive budget, SplatWAM-$\mathcal{T}$ outperforms C3G on every appearance and geometry metric, more than halving its depth error. It also achieves the lowest AbsRel and normal error, even compared with AnySplat, which uses about 180$\times$ as many primitives. This gives us a compact, geometrically grounded state in which SplatWAM-$\mathcal{D}$ can learn dynamics.

\begin{table}[t]
\centering
\scriptsize
\setlength{\tabcolsep}{2.5pt}
\newcommand{\pmv}[1]{{\tiny$\pm$#1}}
\caption{\textbf{Model-free success rate (\%) on OGBench and LIBERO.} Mean $\pm$ standard deviation over three training seeds.}
\label{tab:tokbc}
\begin{tabular*}{\columnwidth}{@{\extracolsep{\fill}}lccccc@{}}
\toprule
& \multicolumn{5}{c}{\textbf{OGBench}} \\
\cmidrule(l){2-6}
Input & Scene & \shortstack{Cube-\\Single} & \shortstack{Cube-\\Double} & \shortstack{Cube-\\Triple} & Avg. \\
\midrule
Pixels & 33.6\pmv{2.4} & 92.9\pmv{3.3} & 50.1\pmv{7.8} & 20.4\pmv{0.4} & 49.3\pmv{2.7} \\
Gaussian Tokens & \textbf{41.6}\pmv{2.8} & \textbf{98.3}\pmv{1.4} & \textbf{74.9}\pmv{3.2} & \textbf{38.9}\pmv{14.5} & \textbf{63.4}\pmv{3.5} \\
\midrule
& \multicolumn{5}{c}{\textbf{LIBERO}} \\
\cmidrule(l){2-6}
Input & Spatial & Object & Goal & Long & Avg. \\
\midrule
Pixels & 70.0\pmv{4.9} & \textbf{87.8}\pmv{3.1} & \textbf{67.6}\pmv{6.1} & 26.5\pmv{2.3} & 63.0\pmv{1.1} \\
Gaussian Tokens & \textbf{80.2}\pmv{0.3} & 84.0\pmv{5.2} & 54.8\pmv{5.9} & \textbf{42.1}\pmv{6.0} & \textbf{65.3}\pmv{4.2} \\
\bottomrule
\end{tabular*}
\end{table}

\subsection{SplatWAM-\texorpdfstring{$\mathcal{D}$}{D} as a simulator}
\label{sec:exp_sim}
\begin{figure*}[t]
\centering
\includegraphics[width=\textwidth]{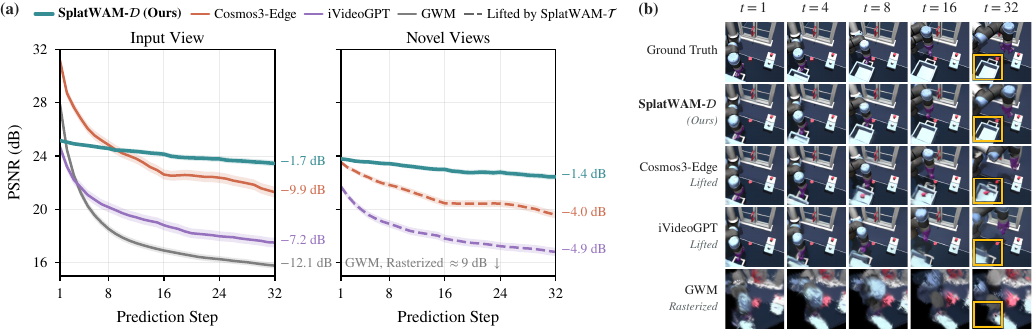}
\caption{\textbf{Action-conditioned rollouts on OGBench Scene.} (a) PSNR versus rollout horizon. (b) Ground truth and predictions from the same context frames and actions. Yellow boxes mark where the baselines drift from the ground truth at $t=32$.}
\label{fig:rollout}
\end{figure*}
\begin{figure*}[t]
\centering
\includegraphics[width=\textwidth]{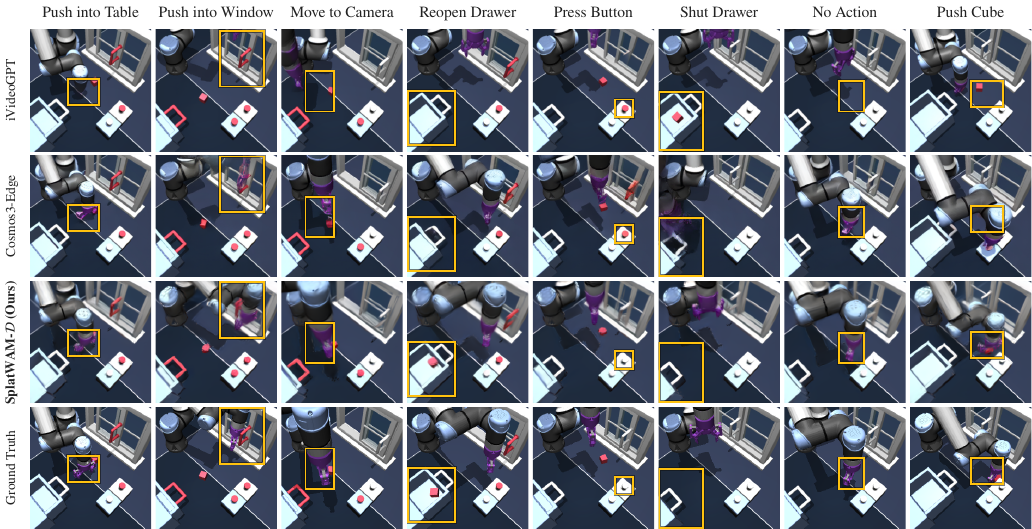}
\caption{\textbf{Counterfactual actions.} Each column uses the same actions and initial states. Success rates over 32 trials are in App.~\ref{sec:app_counterfactual}.}
\label{fig:counterfactual}
\end{figure*}
\begin{table*}[t]
\centering
\scriptsize
\setlength{\tabcolsep}{2.0pt}
\newcommand{\pmv}[1]{{\tiny$\pm$#1}}
\caption{\textbf{Success rate (\%) compared with published methods.} Left: Bold marks the best world-model policy on OGBench, and $\pm$ is the standard deviation over seeds. GCIQL, GCIVL, and HIQL are value-based RL methods provided for reference.
Right: Bold marks the best LIBERO method without robot pretraining. We report the last checkpoint; OpenVLA-OFT reports its best. We also report the standard error over 3 seeds. $^{\dagger}$Hierarchical policy with subgoals. $^{\ddagger}$Uses a wrist camera and proprioception.}
\label{tab:published}
\iflatexml\def\pubpair#1{#1}\def\pubsep{\par}\else\def\pubpair{\makebox[\textwidth][s]}\def\pubsep{\hfill}\fi
\pubpair{%
\begin{tabular}[t]{@{}lccccccc@{}}
\toprule
& & \multicolumn{6}{c}{\textbf{OGBench Scene}} \\
\cmidrule(l){3-8}
Method & RL & Open & \shortstack{Unlock-\\and-Lock} & \shortstack{Rearrange-\\Medium} & \shortstack{Put-in-\\Drawer} & \shortstack{Rearrange-\\Hard} & Overall \\
\midrule
\rowcolor{gray!15}\multicolumn{8}{@{}l}{\textit{Model-Free}} \\
GCBC~\cite{playlmp,gcsl} & \xmark & 59\pmv{7} & 0\pmv{0} & 0\pmv{0} & 2\pmv{1} & 0\pmv{0} & 12\pmv{2} \\
GCIQL~\cite{iql} & \cmark & 56\pmv{4} & 1\pmv{1} & 0\pmv{0} & 3\pmv{4} & 0\pmv{0} & 12\pmv{2} \\
GCIVL~\cite{hiql} & \cmark & 84\pmv{4} & 24\pmv{8} & 16\pmv{8} & 0\pmv{0} & 0\pmv{0} & 25\pmv{3} \\
HIQL$^{\dagger}$~\cite{hiql} & \cmark & 80\pmv{6} & 81\pmv{7} & 61\pmv{11} & 20\pmv{8} & 3\pmv{2} & 49\pmv{4} \\
\midrule
\rowcolor{gray!15}\multicolumn{8}{@{}l}{\textit{World Model}} \\
LPWM~\cite{lpwm} & \xmark & \textbf{100}\pmv{0} & 6\pmv{9} & 89\pmv{9} & \textbf{3}\pmv{5} & 0\pmv{0} & 40\pmv{1} \\
\textbf{SplatWAM-$\mathcal{P}$ (Ours)} & \xmark & \textbf{100}\pmv{0} & \textbf{67}\pmv{13} & \textbf{91}\pmv{3} & \textbf{3}\pmv{3} & 0\pmv{0} & \textbf{52}\pmv{2} \\
\bottomrule
\end{tabular}\pubsep
\begin{tabular}[t]{@{}lcccccc@{}}
\toprule
& & \multicolumn{5}{c}{\textbf{LIBERO}} \\
\cmidrule(l){3-7}
Method & Size & Spatial & Object & Goal & Long & Avg. \\
\midrule
\rowcolor{gray!15}\multicolumn{7}{@{}l}{\textit{Robot Pretraining}} \\
Octo~\cite{octo} & ${\sim}$90M & 78.9\pmv{1.0} & 85.7\pmv{0.9} & 84.6\pmv{0.9} & 51.1\pmv{1.3} & 75.1\pmv{0.6} \\
OpenVLA~\cite{openvla} & ${\sim}$7B & 84.7\pmv{0.9} & 88.4\pmv{0.8} & 79.2\pmv{1.0} & 53.7\pmv{1.3} & 76.5\pmv{0.6} \\
OpenVLA-OFT$^{\ddagger}$~\cite{openvlaoft} & ${\sim}$7B & 97.6 & 98.4 & 97.9 & 94.5 & 97.1 \\
\midrule
\rowcolor{gray!15}\multicolumn{7}{@{}l}{\textit{No Robot Pretraining}} \\
Diffusion Policy~\cite{diffusionpolicy} & -- & 78.3\pmv{1.1} & 92.5\pmv{0.7} & 68.3\pmv{1.2} & 50.5\pmv{1.3} & 72.4\pmv{0.7} \\
WorldVLA~\cite{worldvla} & ${\sim}$7B & 85.6 & 89.0 & 82.6 & 59.0 & 79.1 \\
OpenVLA-OFT$^{\ddagger}$~\cite{openvlaoft} & ${\sim}$7B & \textbf{94.3} & 95.2 & 91.7 & \textbf{86.5} & 91.9 \\
\textbf{SplatWAM-$\mathcal{P}$ (Ours)} & \textbf{${\sim}$1B} & 92.6\pmv{0.7} & \textbf{99.0}\pmv{0.3} & \textbf{92.7}\pmv{0.7} & 84.7\pmv{0.9} & \textbf{92.3}\pmv{0.3} \\
\bottomrule
\end{tabular}}
\end{table*}
\begin{table}[t]
\centering
\footnotesize
\setlength{\tabcolsep}{3pt}
\caption{\textbf{Dynamics ablation on OGBench Scene.} Cells report results at horizons $H=16$/$H=32$.}
\label{tab:ablation}
\begin{tabular}{@{}lcccc@{}}
\toprule
& PSNR$\uparrow$ & SSIM$\uparrow$ & LPIPS$\downarrow$ & FVD$\downarrow$ \\
\midrule
\rowcolor{gray!15}\multicolumn{5}{@{}l}{\textit{Input Views}} \\
\quad Full & \textbf{24.57/24.15} & \textbf{.883/.874} & \textbf{.085/.088} & 173/338 \\
\quad $-$3D PE & 24.39/23.89 & .878/.868 & .086/.090 & 170/331 \\
\quad $-$log-cov & 24.47/23.89 & .880/.868 & \textbf{.085}/.090 & \textbf{168/324} \\
\quad $-$render & 23.17/22.71 & .851/.841 & .100/.104 & 209/420 \\
\quad $-$log-cov, $-$render & 23.55/23.15 & .860/.851 & .096/.100 & 202/413 \\
\midrule
\rowcolor{gray!15}\multicolumn{5}{@{}l}{\textit{Novel Views}} \\
\quad Full & \textbf{23.37/23.02} & \textbf{.871/.863} & \textbf{.099/.102} & 181/\textbf{313} \\
\quad $-$3D PE & 23.29/22.87 & .867/.857 & .100/.104 & \textbf{176}/316 \\
\quad $-$log-cov & 23.28/22.80 & .868/.856 & \textbf{.099}/.104 & 181/318 \\
\quad $-$render & 22.31/21.92 & .842/.832 & .114/.119 & 221/417 \\
\quad $-$log-cov, $-$render & 22.63/22.28 & .850/.841 & .111/.115 & 224/405 \\
\bottomrule
\end{tabular}
\end{table}
We next evaluate open-loop, action-conditioned rollouts on OGBench Scene against iVideoGPT~\cite{ivideogpt}, GWM~\cite{gwm}, and Cosmos3-Edge~\cite{cosmos3}, restricting the comparison to world models that run on a single consumer GPU, as real-world deployment requires. We evaluate at $H=16$ and $H=32$, with the latter testing extrapolation beyond SplatWAM-$\mathcal{D}$'s training horizon. Table~\ref{tab:rollout} reports context-view prediction and novel-view rendering at three held-out views.
Because iVideoGPT and Cosmos3-Edge predict only context views, we lift their outputs post-hoc using SplatWAM-$\mathcal{T}$ for a fair comparison.
In contrast, GWM and SplatWAM-$\mathcal{D}$ predict Gaussian states that can be rasterized and compared directly.
We measure appearance quality with PSNR, SSIM, and LPIPS, clip-level quality with FVD, and cross-view consistency with MEt3R~\cite{met3r}.
End-to-end inference speed is measured on a single RTX 5090 at batch size one, including lifting where required.
Evaluation and metric details are in App.~\ref{sec:app_wm_eval}.

Cosmos3-Edge, pretrained at scale, achieves the highest perceptual quality when constrained to the provided context views. However, SplatWAM-$\mathcal{D}$ has the strongest novel-view appearance and rollout quality while maintaining comparable cross-view consistency and substantially higher rendering throughput. Its rollouts remain temporally coherent as the horizon grows (Fig.~\ref{fig:rollout}), while PSNR degrades little beyond the training horizon. Because each predicted state is explicitly rasterizable, SplatWAM can render arbitrary viewpoints and output resolutions without additional generative sampling, incurring only the cost of rasterization for each new view.

To assess action fidelity beyond logged rollouts, we evaluate counterfactual actions from held-out OGBench Scene states (Fig.~\ref{fig:counterfactual}). We observe that video models tend to create interpenetration scenarios, with iVideoGPT pressing the end-effector through the table and Cosmos3-Edge merging it with the window. SplatWAM-$\mathcal{D}$ stops the gripper at both surfaces in every trial. We also find these video models follow actions less faithfully. They often miss button presses, and iVideoGPT fails to shut the drawer, while SplatWAM-$\mathcal{D}$ carries out both actions in every trial. The largest gap is object permanence. When the arm shuts the cube inside the drawer and later reopens it, SplatWAM-$\mathcal{D}$ brings the cube back in every trial, whereas the video models often lose it. These results suggest that SplatWAM-$\mathcal{D}$'s improved visual fidelity and geometric consistency are accompanied by more faithful counterfactual action predictions, particularly for contact interactions and object permanence. Action marginalization~\cite{wmhallucination} remains a limitation even for SplatWAM-$\mathcal{D}$, and cube pushing, which is rare in the training data, remains the hardest test for all models.

\subsection{SplatWAM-\texorpdfstring{$\mathcal{P}$}{P} for control}
\label{sec:exp_control}
We evaluate the Gaussian representation for model-free control and the performance of a policy that reuses the predictive world-model backbone. All policies use two context views. Detailed protocols are in App.~\ref{sec:app_pol_protocol}.

To compare pixel and Gaussian representations for control, we train two model-free policies with matched observation views
and output embedding dimensions, using the same flow-matching action head. One policy encodes pixels with an IMPALA CNN~\cite{impala}, and the other pools the Gaussian tokens of the current frame with a Set Transformer~\cite{settransformer}. Gaussian tokens outperform pixels on six of the eight task suites (Table~\ref{tab:tokbc}), primarily on tasks that depend on 3D spatial relations among objects, such as rearranging cubes or manipulating objects in varying layouts. We attribute these gains to SplatWAM-$\mathcal{T}$'s learned geometric representation, rather than its VGGT features alone (App.~\ref{sec:app_vggt}).
Pixels lead only on Object and Goal, two cluttered suites that require precise grasps of small objects. With more context views, tokens close the gap (App.~\ref{sec:app_repcontrol}).

We next reuse SplatWAM-$\mathcal{D}$ as the policy backbone and jointly optimize it with SplatWAM-$\mathcal{P}$ (\S\ref{sec:policy}). This improves over the model-free Gaussian policy on every LIBERO suite, particularly Goal and Long, consistent with temporal context mitigating sparse-view occlusion. Against published results (Table~\ref{tab:published}), it outperforms LPWM on OGBench Scene, driven by Unlock-and-Lock, a long-horizon task requiring sequential execution of six atomic behaviors. It also stays competitive with HIQL without learning a value function or using subgoals for horizon reduction. On LIBERO, without robot pretraining, SplatWAM-$\mathcal{P}$ outperforms the larger WorldVLA and OpenVLA. It also matches OpenVLA-OFT trained without robot pretraining, despite fewer parameters and no wrist camera or proprioception.

\subsection{Ablations}
\label{sec:exp_ablation}

Table~\ref{tab:ablation} ablates the 3D positional embedding (3D PE), rendering loss $\mathcal{L}_{\mathrm{render}}$ (render), and log-Euclidean covariance reparameterization (log-cov) under the same two-view inputs and evaluation protocol as Table~\ref{tab:rollout}. Removing rendering supervision causes the largest degradation in both input and novel-view quality. Removing log-cov slightly improves input-view FVD but reduces PSNR and SSIM. The reparameterization also removes quaternion-scale ambiguities while preserving positive-definite covariance matrices. Finally, 3D PE has little effect on final performance but accelerates convergence by roughly one-third (App.~\ref{sec:app_wm_impl}). Additional ablations (App.~\ref{sec:app_joint}) show that joint training degrades rollout quality, a gradient interference effect also observed in VLAs~\cite{knowinsulate}. Blocking action-head gradients preserves predictive quality but slows policy learning, consistent with~\cite{gradrouting}. Adding $\mathcal{L}_{\mathrm{render}}$ back during joint training ($\mathcal{L}_{\pi}+\mathcal{L}_{\mathrm{dyn}}$) instead mitigates this degradation without sacrificing policy success.

    \section{Conclusion}
\label{sec:conclusion}
We introduced SplatWAM, a generative world action model that learns action-conditioned dynamics directly over compact 3D Gaussian states. By predicting explicit, rasterizable scene representations rather than compressed latents, SplatWAM supports efficient novel-view rollouts with a shared geometric representation across views. Policies built on this state outperform pixel-based policies and remain competitive with fine-tuned VLAs, without large-scale robot pretraining. Future work will explore permutation-equivariant dynamics that treat the Gaussians as an unordered set, richer semantic Gaussian attributes for editable world models, and reinforcement learning in imagined 3D rollouts.

\FloatBarrier
{
    \small
    \bibliographystyle{ieeenat_fullname}
    \bibliography{main}
}

\clearpage
\raggedbottom
\twocolumn[\centering\Large\textbf{SplatWAM: Splatting World Action Models for Real-Time 3D Imagination}\\\vspace{0.5em}Appendix\\\vspace{1.0em}]
\appendix
\counterwithin*{table}{section}
\counterwithin*{figure}{section}
\renewcommand{\thetable}{\thesection\arabic{table}}
\renewcommand{\thefigure}{\thesection\arabic{figure}}
\renewcommand{\theHtable}{app.\thesection.\arabic{table}}
\renewcommand{\theHfigure}{app.\thesection.\arabic{figure}}
\setcounter{table}{0}
\setcounter{figure}{0}

\section{Tokenizer (SplatWAM-\texorpdfstring{$\mathcal{T}$}{T})}
\label{sec:app_tok}

\subsection{Implementation}
\label{sec:app_tok_impl}

SplatWAM-$\mathcal{T}$ is initialized from C3G~\cite{c3g} and fine-tuned on our data, including its VGGT backbone, using the hyperparameters in Table~\ref{tab:app_tok_hparams}.

\begin{table}[!htb]
\centering
\footnotesize
\caption{\textbf{SplatWAM-$\mathcal{T}$ hyperparameters.}}
\label{tab:app_tok_hparams}
\begin{tabular}{@{}ll@{}}
\toprule
Hyperparameter & Value \\
\midrule
\rowcolor{gray!15}\multicolumn{2}{@{}l}{\textit{Architecture}} \\
Gaussian tokens $N$ & 2048 \\
\midrule
\rowcolor{gray!15}\multicolumn{2}{@{}l}{\textit{Objective}} \\
Perceptual weight $\lambda_p$ & 0.05 \\
Depth weight $\lambda_d$ & 0.01 \\
Normal weight $\lambda_n$ & 0.001 \\
Coverage weight $\lambda_\alpha$ & 0.02 \\
Anisotropy weight $\lambda_{\mathrm{aniso}}$ & 0.01 \\
Effective-rank weight $\lambda_{\mathrm{erank}}$ & 0.01 \\
Maximum aspect ratio $r$ & 10 \\
Effective-rank offset $\delta$ & $10^{-5}$ \\
\midrule
\rowcolor{gray!15}\multicolumn{2}{@{}l}{\textit{Optimization}} \\
Optimizer & AdamW \\
Learning rate & $10^{-5}$ \\
Training steps & 30K \\
\bottomrule
\end{tabular}
\end{table}

Before rasterization, we normalize each output quaternion, apply softplus scaled by $10^{-3}$ to ensure positive scales, and map opacity through a sigmoid. We represent colors using degree-0 spherical harmonics and denote the positive scales by $\varsigma_i$. SplatWAM-$\mathcal{D}$ predicts unconstrained position, opacity, and color parameters, together with log-Euclidean covariance coordinates $\nu_i$ (\S\ref{sec:gaussiandyn}), which are converted to valid Gaussian parameters before rendering.

We use a log-space variant of the PhysGaussian anisotropy regularizer~\cite{physgaussian}:
\begin{equation}
\mathcal{L}_{\mathrm{aniso}} = \mathbb{E}_i\big[\max\big(0,\ \log(\varsigma_{i,\max}/\varsigma_{i,\min}) - \log r\big)\big],
\end{equation}
where $\varsigma_{i,\max}$ and $\varsigma_{i,\min}$ are the largest and smallest scales of primitive $i$ and $r$ is the maximum allowed aspect ratio. The effective-rank term is
\begin{equation}
\mathcal{L}_{\mathrm{erank}} = \mathbb{E}_i\big[\max\big(0,\ -\log(\operatorname{erank}(\varsigma_i) - 1 + \delta)\big)\big],
\end{equation}
where $\operatorname{erank}(\varsigma_i) = \exp\bigl(-\sum_k \sigma_k \log \sigma_k\bigr)$ with $\sigma_k = \varsigma_{i,k}^2 / \sum_j \varsigma_{i,j}^2$ is the effective rank of the covariance~\cite{erankgs}.

\begin{figure}[t]
\centering
\includegraphics[width=\columnwidth]{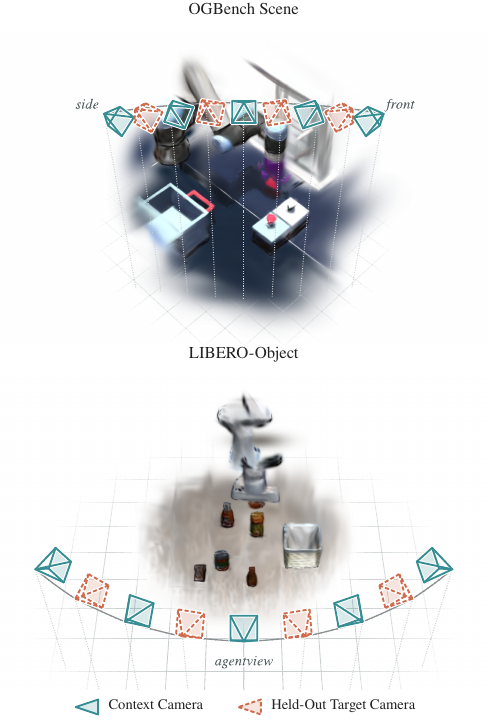}
\caption{\textbf{Camera setup.} Nine cameras on a fixed arc: five context cameras (teal) and four held-out target cameras (coral) midway between them.}
\label{fig:app_rig}
\end{figure}
\begin{figure*}[t]
\centering
\includegraphics[width=0.90\textwidth]{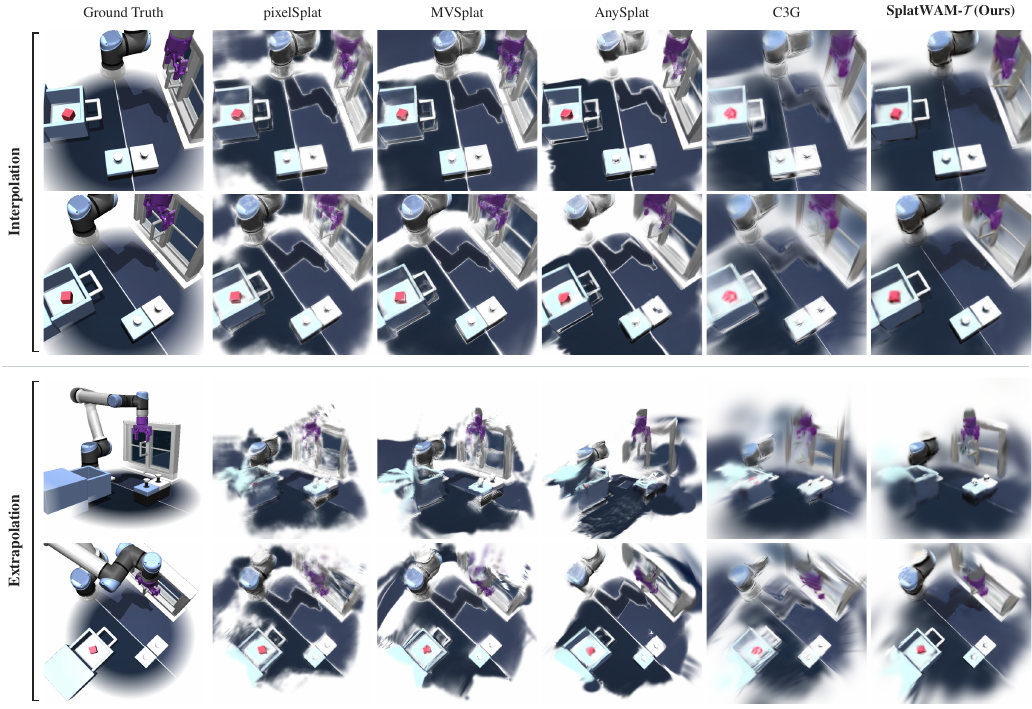}
\caption{\textbf{Novel-view interpolation and extrapolation.} All methods use the same two context views as in Table~\ref{tab:ffgs}. We render two held-out viewpoints between the context cameras (interpolation) and two off-rig viewpoints, one low and one overhead (extrapolation). SplatWAM-$\mathcal{T}$ preserves scene structure with 2K primitives.}
\label{fig:app_qual}
\end{figure*}

\subsection{Evaluation protocol}
\label{sec:app_tok_eval}

We evaluate SplatWAM-$\mathcal{T}$ on 30 held-out scenes, each a random state from a different official OGBench Scene validation episode, using two $224\times224$ context views. Novel-view metrics are evaluated at three viewpoints not used as inputs: one of the unused context-camera positions and two held-out target-camera positions (Fig.~\ref{fig:app_rig}). All trainable baselines are fine-tuned from released checkpoints with the same training schedule, photometric objective, scenes, and target views. We align predicted depth to simulator depth by a per-image least-squares scale and shift over valid pixels. Surface-normal error is the mean angular difference between normals derived from the aligned depth and simulator depth, measured at that context-camera position in Table~\ref{tab:ffgs} and averaged over all three viewpoints in Table~\ref{tab:app_suites}.

\subsection{Additional results}
\label{sec:app_repr}

\begin{table}[!t]
\centering
\footnotesize
\setlength{\tabcolsep}{1.8pt}
\caption{\textbf{Novel-view synthesis and geometry of SplatWAM-$\mathcal{T}$ across OGBench and LIBERO.} LIBERO-Long is excluded from tokenizer pretraining to evaluate representation transfer.}
\label{tab:app_suites}
\begin{tabular}{@{}lcccccc@{}}
\toprule
Suite & PSNR$\uparrow$ & SSIM$\uparrow$ & LPIPS$\downarrow$ & AbsRel$\downarrow$ & $\delta_1\uparrow$ & Normal\,($^\circ$)$\downarrow$ \\
\midrule
\rowcolor{gray!15}\multicolumn{7}{@{}l}{\textit{OGBench}} \\
\quad Scene & 23.00 & .837 & .149 & .055 & 95.0 & 38.0 \\
\quad Cube-Single & 27.57 & .919 & .074 & .125 & 87.7 & 40.8 \\
\quad Cube-Double & 27.78 & .921 & .074 & .120 & 89.7 & 40.4 \\
\quad Cube-Triple & 27.21 & .917 & .079 & .114 & 90.7 & 40.1 \\
\midrule
\rowcolor{gray!15}\multicolumn{7}{@{}l}{\textit{LIBERO}} \\
\quad Spatial & 24.45 & .817 & .238 & .035 & 98.8 & 30.6 \\
\quad Object & 24.71 & .765 & .271 & .036 & 98.7 & 31.0 \\
\quad Goal & 23.98 & .809 & .229 & .040 & 97.6 & 32.9 \\
\quad 90 & 21.94 & .725 & .298 & .046 & 97.4 & 35.8 \\
\quad Long (OOD) & 21.80 & .734 & .297 & .046 & 97.1 & 36.9 \\
\bottomrule
\end{tabular}
\end{table}

\begin{figure*}[p]
\renewcommand{\thefigure}{A3}%
\centering
\includegraphics[width=\textwidth]{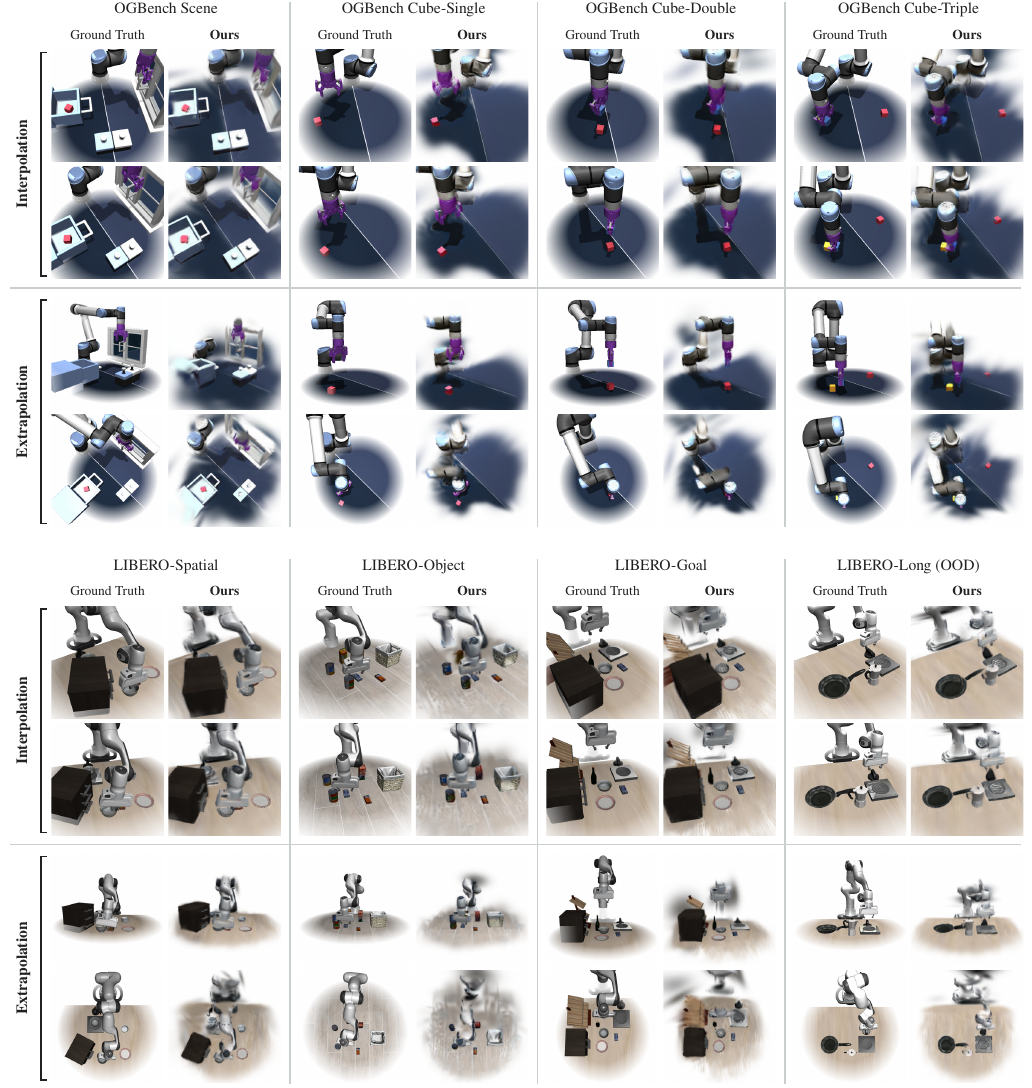}
\caption{\textbf{Qualitative results of SplatWAM-$\mathcal{T}$.} Novel-view interpolation and extrapolation across OGBench and LIBERO. Camera viewpoints follow the setup in Figs.~\ref{fig:app_rig} and~\ref{fig:app_qual}.}
\label{fig:app_qual_suites}
\end{figure*}

Table~\ref{tab:app_suites} reports SplatWAM-$\mathcal{T}$ across OGBench and LIBERO. Fig.~\ref{fig:app_qual} compares the methods in Table~\ref{tab:ffgs} under view interpolation and extrapolation. Pixel-aligned methods (pixelSplat, MVSplat) produce sharper interpolation results, as reflected in their higher appearance scores, but exhibit floating artifacts at extrapolated viewpoints. C3G and SplatWAM-$\mathcal{T}$ better preserve the scene layout with 2K primitives. However, C3G produces blurry, semi-transparent renderings of the arm and window, whereas SplatWAM-$\mathcal{T}$ preserves their structure and opacity, consistent with its lower depth and normal errors. Both methods blur regions not visible in the context views. Fig.~\ref{fig:app_qual_suites} shows additional interpolation and extrapolation results across OGBench and LIBERO.

\section{World model (SplatWAM-\texorpdfstring{$\mathcal{D}$}{D})}
\label{sec:app_wm}

\subsection{Shortcut forcing}
\label{sec:app_sf}

SplatWAM-$\mathcal{D}$ (\S\ref{sec:dynamics}) applies the shortcut-forcing objective of Dreamer 4~\cite{dreamer4} to the log-Euclidean Gaussian states of Eq.~\eqref{eq:df}:
{\setlength{\multlinegap}{0pt}
\begin{multline}
\mathcal{L}_{\mathrm{SF}}=\\
\mathbb{E}\!\left[
\omega(\tau_t)
\begin{cases}
\|\hat{\tilde{s}}_t-\tilde{s}_t\|_2^2, & d_t=d_{\min},\\[2pt]
(1-\tau_t)^2\left\|\hat{u}_t-\operatorname{sg}\!\left(\frac{u'_t+u''_t}{2}\right)\right\|_2^2, & d_t>d_{\min},
\end{cases}
\right].
\end{multline}}
Here $d_t$ is the shortcut step size in signal level $\tau_t$, and $\hat{u}_t=(\hat{\tilde{s}}_t-z_t)/(1-\tau_t)$ is the velocity implied by the predicted clean state. The velocities $u'_t$ and $u''_t$ are computed over two successive half-steps of size $d_t/2$, with the second starting from the intermediate state reached by the first. The operator $\operatorname{sg}$ stops gradients, and $\omega(\tau_t)$ increases linearly with the signal level. At the finest step, the objective reduces to clean-state prediction; larger steps match the average velocity of the two half-steps~\cite{shortcut}.

\subsection{Implementation}
\label{sec:app_wm_impl}

\begin{figure}[!htb]
\centering
\includegraphics[width=\linewidth]{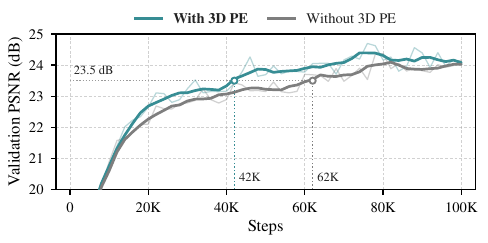}
\caption{\textbf{Effect of the 3D positional embedding on convergence.} Both variants reach similar final validation PSNR, but the 3D positional embedding reduces the number of steps required to reach 23.5\,dB from 62K to 42K.}
\label{fig:app_xyzpe}
\end{figure}

SplatWAM-$\mathcal{D}$ uses register tokens and temporal attention in every fourth transformer layer. Table~\ref{tab:app_wm_hparams} lists the hyperparameters shared across benchmarks. We optimize with Muon~\cite{muon}, which reaches AdamW's final validation error after approximately 20K steps on OGBench Scene and achieves 40\% lower error after 100K steps under the same settings. The 3D positional embedding also accelerates convergence, reaching 23.5\,dB validation PSNR in 42K steps instead of 62K, although the difference narrows by 100K steps (Fig.~\ref{fig:app_xyzpe}).

For action dropout, we replace all actions in 10\% of training sequences with a learned null token, following classifier-free guidance~\cite{cfg}. The policy pass uses the same null token. At inference, each future Gaussian state is generated with a single shortcut step ($K=1$). Observed states use $\tau=1-d_{\min}$, while generated states are re-noised to $\tau_{\mathrm{ctx}}$ when reused as context. For rollouts longer than the training window, the context advances by eight frames, retaining the most recent 9--16 states.

\begin{table}[!htb]
\centering
\footnotesize
\caption{\textbf{SplatWAM-$\mathcal{D}$ hyperparameters.}}
\label{tab:app_wm_hparams}
\begin{tabular}{@{}ll@{}}
\toprule
Hyperparameter & Value \\
\midrule
\rowcolor{gray!15}\multicolumn{2}{@{}l}{\textit{Architecture}} \\
Layers & 12 \\
Width & 512 \\
Attention heads & 8 \\
Register tokens & 4 \\
Parameters & 54M \\
Window length $T$ & 17 \\
\midrule
\rowcolor{gray!15}\multicolumn{2}{@{}l}{\textit{Objective}} \\
Finest step size $d_{\min}$ & 1/64 \\
Batch fraction with $d_t>d_{\min}$ (from step 5K) & 0.25 \\
Loss weight $\omega(\tau)$ & $0.9\tau+0.1$ \\
Rendering weight $\lambda_{\mathrm{render}}$ & 0.2 \\
Action dropout & 0.1 \\
\midrule
\rowcolor{gray!15}\multicolumn{2}{@{}l}{\textit{Optimization}} \\
Optimizer & Muon \\
Learning rate & $3\times10^{-4}$ \\
Warmup & 2K steps \\
LR schedule & cosine annealing \\
Weight decay & $10^{-3}$ \\
Gradient clipping & 1.0 \\
Batch size & 16 \\
Training steps & 100K \\
EMA decay & 0.999 \\
\midrule
\rowcolor{gray!15}\multicolumn{2}{@{}l}{\textit{Inference}} \\
Denoising steps per frame $K$ & 1 \\
Context window stride & 8 frames \\
Context signal level $\tau_{\mathrm{ctx}}$ & 0.8 \\
\bottomrule
\end{tabular}
\end{table}

\subsection{Evaluation protocol}
\label{sec:app_wm_eval}

\begin{figure*}[p]
\renewcommand{\thefigure}{B2}%
\centering
\includegraphics[width=\textwidth]{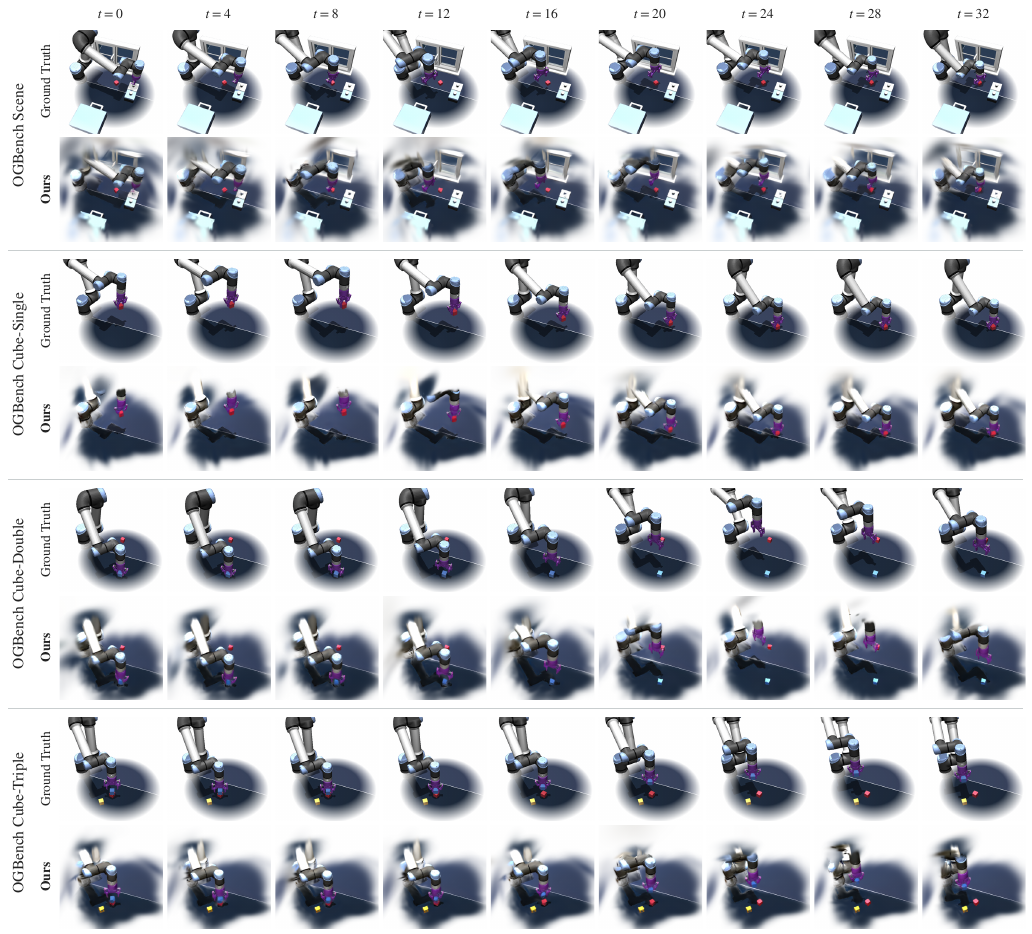}
\caption{\textbf{SplatWAM-$\mathcal{D}$'s rollouts on OGBench.} We predict 32 steps from two context views and render the Gaussian states from a novel viewpoint.}
\label{fig:app_wm_rollouts}
\end{figure*}

\begin{figure*}[p]
\renewcommand{\thefigure}{B3}%
\centering
\includegraphics[width=\textwidth]{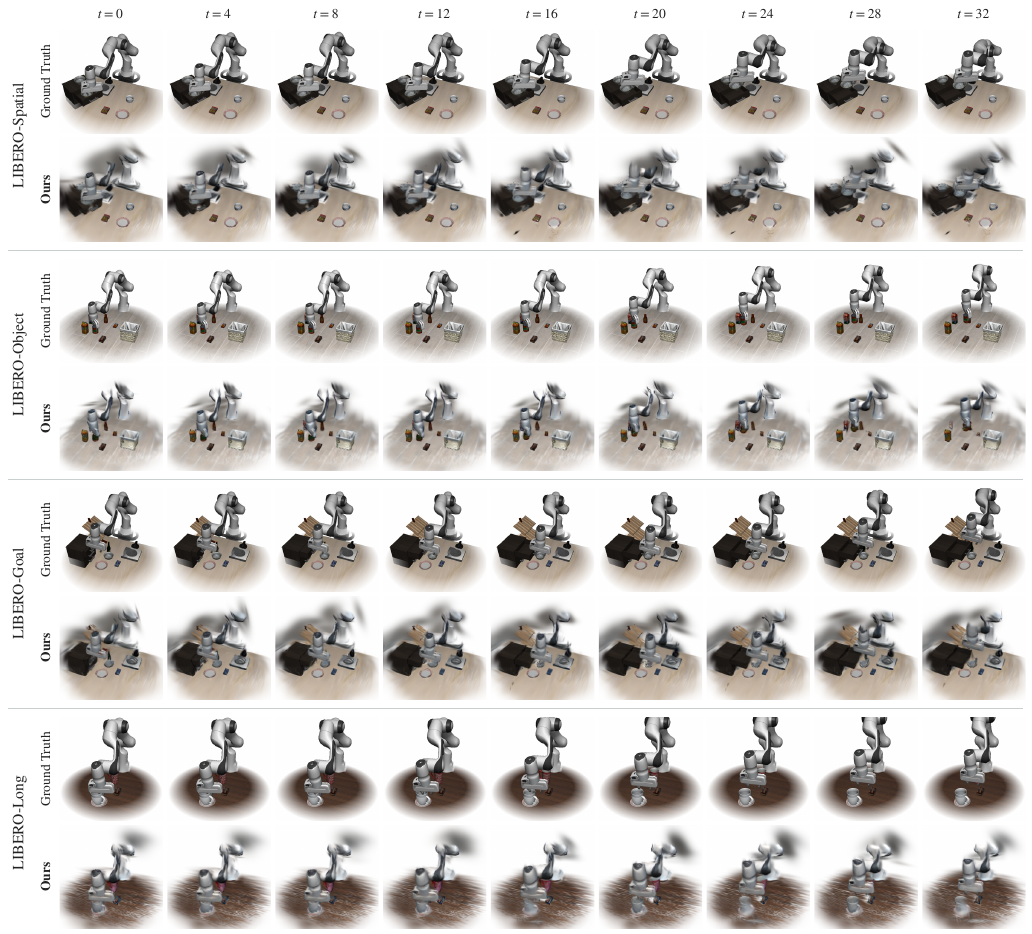}
\caption{\textbf{SplatWAM-$\mathcal{D}$'s rollouts on LIBERO.} Evaluation follows the same setup as Fig.~\ref{fig:app_wm_rollouts}.}
\label{fig:app_wm_rollouts_libero}
\end{figure*}

For Table~\ref{tab:rollout}, we train or fine-tune all models on the same OGBench trajectories until validation loss saturates. We evaluate 400 windows sampled from 100 held-out validation episodes. Each rollout starts from two context views and follows the logged actions without observing future frames. Models retain their supported input resolutions, while predictions and ground truth are evaluated at $128\times128$, the highest common output resolution. Figs.~\ref{fig:app_wm_rollouts} and~\ref{fig:app_wm_rollouts_libero} show additional Gaussian-state rollouts on OGBench and LIBERO.

\subsection{Counterfactual evaluation}
\label{sec:app_counterfactual}

\begin{table}[!t]
\centering
\footnotesize
\setlength{\tabcolsep}{3pt}
\caption{\textbf{Counterfactual success rate (\%).} Each task is evaluated on 32 held-out OGBench Scene states using the same prescribed action sequences. Success is determined from predicted frames using task-specific criteria described in \S\ref{sec:app_counterfactual}.}
\label{tab:counterfactual}
\begin{tabular}{@{}lrrr@{}}
\toprule
Task & iVideoGPT & Cosmos3-Edge & SplatWAM-$\mathcal{D}$ \\
\midrule
Push into Table  & 72 \scriptsize{(23/32)} & \textbf{100} \scriptsize{(32/32)} & \textbf{100} \scriptsize{(32/32)} \\
Push into Window & 19 \scriptsize{(6/32)} & 62 \scriptsize{(20/32)} & \textbf{100} \scriptsize{(32/32)} \\
Move to Camera   & 38 \scriptsize{(12/32)} & 88 \scriptsize{(28/32)} & \textbf{91} \scriptsize{(29/32)} \\
Reopen Drawer    & 3 \scriptsize{(1/32)} & 38 \scriptsize{(12/32)} & \textbf{100} \scriptsize{(32/32)} \\
Press Button     & 22 \scriptsize{(7/32)} & 69 \scriptsize{(22/32)} & \textbf{100} \scriptsize{(32/32)} \\
Shut Drawer      & 19 \scriptsize{(6/32)} & 97 \scriptsize{(31/32)} & \textbf{100} \scriptsize{(32/32)} \\
No Action        & 66 \scriptsize{(21/32)} & 69 \scriptsize{(22/32)} & \textbf{75} \scriptsize{(24/32)} \\
Push Cube        & 0 \scriptsize{(0/32)} & 0 \scriptsize{(0/32)} & \textbf{41} \scriptsize{(13/32)} \\
\bottomrule
\end{tabular}
\end{table}

Table~\ref{tab:counterfactual} reports counterfactual action success rates evaluated from predicted frames (Fig.~\ref{fig:counterfactual}), rather than predicted simulator states. Each task uses 32 held-out OGBench Scene states, selected before evaluation to ensure that the expected outcome is visible from the camera. All models receive the same prescribed action sequences, converted to their respective action formats when necessary. For Cosmos3-Edge, we convert commands to free-space end-effector displacements. Pushes still proceed into contact for all models.

Push, move-to-camera, and no-action tasks use scripted commands with a zero gripper command. Pushes continue for 24 steps into the table and 60 steps into the window after contact. Button and drawer tasks use OGBench's scripted data-collection policy. The gripper starts above the button; for Reopen Drawer, the policy closes the drawer with a cube inside, waits, and reopens it.

We score success from the predicted frames using task-specific criteria: the gripper must stop at the table, remain in front of the window, or increase in apparent size by at least 10\% when moving toward the camera; the button must toggle, the drawer must remain shut, or the gripper must stay within 20 pixels under no action. For cube pushing, the predicted cube displacement must reach at least half the ground-truth cube displacement. Reopening the drawer requires the cube to disappear and reappear; Cosmos3-Edge and iVideoGPT hide it in only 21 and 5 of 32 trials, respectively. These comparisons measure counterfactual behavior at the system level and do not isolate representation effects from learned dynamics.

\section{Policy (SplatWAM-\texorpdfstring{$\mathcal{P}$}{P})}
\label{sec:app_policy}

\subsection{Benchmarks and protocol}
\label{sec:app_pol_protocol}

On OGBench, we train goal-conditioned policies on the official play datasets with hindsight goal relabeling~\cite{her,gcsl}, following learning from play~\cite{playlmp}. For each $T$-frame context, we sample a goal $\Delta$ frames after the window, with $\Delta\sim\mathcal{U}\{1,\dots,\Delta_{\max}\}$, and supervise the corresponding logged action chunk. We evaluate five official goals per environment with binary success, 50 closed-loop rollouts per goal, and the official step budgets: 750 for Scene and 200, 500, and 1000 for single, double, and triple Cube.
On LIBERO, we train a language-conditioned policy separately on each suite's demonstrations. Following OpenVLA~\cite{openvla}, we evaluate 50 initial states for each of 10 tasks (500 episodes per seed, three seeds) under its per-suite step budgets. Policies predict 16 actions, execute 8, and replan.

\subsection{Model-free policies}

\begin{figure*}[t]
\centering
\includegraphics[width=\textwidth]{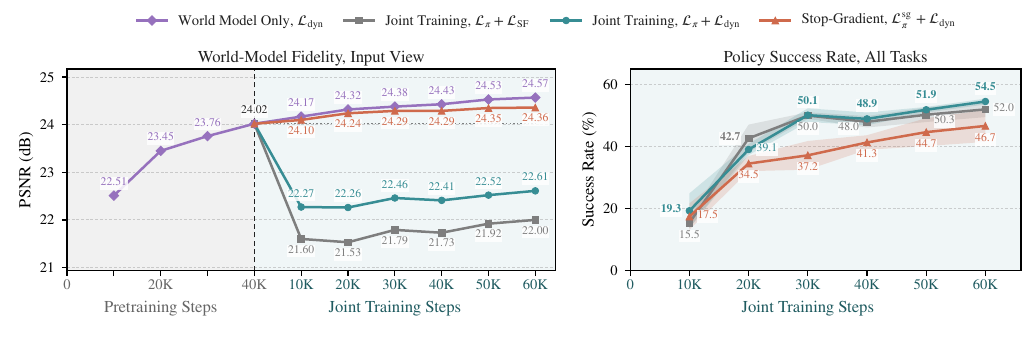}
\caption{\textbf{Joint training on OGBench Scene.} Left: 16-step input-view rollout PSNR during world-model pretraining and subsequent joint training. All variants start from the same 40K checkpoint. Right: policy success over 60K joint-training steps; mean $\pm$ standard deviation over three training seeds. Stop-gradient blocks policy gradients from updating the shared world-model backbone.}
\label{fig:app_joint}
\end{figure*}

The model-free policies in Table~\ref{tab:tokbc} use the same two RGB observations, a 512-dimensional embedding, and a shared flow-matching action-head architecture (Table~\ref{tab:app_mf_hparams}). They differ in observation encoding and augmentation. The pixel policy applies the small IMPALA CNN~\cite{impala} used by OGBench~\cite{ogbench} to channel-stacked views and, on OGBench, the goal image. The token policy applies a Set Transformer~\cite{settransformer} to Gaussian tokens from frozen SplatWAM-$\mathcal{T}$, adding goal-state tokens on OGBench. It pools the set into eight 64-dimensional vectors, concatenated into the embedding. On LIBERO, we project a pooled CLIP~\cite{clip} instruction embedding and concatenate it with the CNN features for the pixel policy, or append it as one token to the Gaussian set for the token policy.

\begin{table}[!htb]
\centering
\footnotesize
\setlength{\tabcolsep}{3pt}
\caption{\textbf{Model-free policy hyperparameters.} All policies use the same two RGB observations. Input shapes refer to the tensors passed to each policy encoder, after any representation-specific preprocessing. Shared hyperparameters apply to all three policies.}
\label{tab:app_mf_hparams}
\begin{tabular}{@{}lccc@{}}
\toprule
Hyperparameter & Pixels & \shortstack{VGGT\\Features} & \shortstack{Gaussian\\Tokens} \\
\midrule
\rowcolor{gray!15}\multicolumn{4}{@{}l}{\textit{Architecture}} \\
Policy input shape & $2\times64\times64\times3$ & $512\times48$ & $2048\times14$ \\
Encoder & IMPALA CNN & \multicolumn{2}{c}{Set Transformer} \\
Encoder width & 16, 32, 32 channels & \multicolumn{2}{c}{64, 4 heads} \\
Inducing points & -- & \multicolumn{2}{c}{32} \\
Pooled features & $32\times8\times8$ & \multicolumn{2}{c}{$8\times64$} \\
Embedding & \multicolumn{3}{c}{512} \\
Window length $T$ & \multicolumn{3}{c}{1 (current frame)} \\
Action head & \multicolumn{3}{c}{flow matching, MLP $3\times512$} \\
\midrule
\rowcolor{gray!15}\multicolumn{4}{@{}l}{\textit{Objective}} \\
Action chunk $k$ & \multicolumn{3}{c}{1 (OGBench), 16 (LIBERO)} \\
Goal horizon $\Delta_{\max}$ & \multicolumn{3}{c}{100 (OGBench)} \\
\midrule
\rowcolor{gray!15}\multicolumn{4}{@{}l}{\textit{Optimization}} \\
Augmentation & random crop ($p=0.5$) & \multicolumn{2}{c}{none} \\
Optimizer & \multicolumn{3}{c}{Adam} \\
Learning rate & \multicolumn{3}{c}{$3\times10^{-4}$} \\
Batch size & \multicolumn{3}{c}{256} \\
Training steps & \multicolumn{3}{c}{500K (OGBench), 100K (LIBERO)} \\
\midrule
\rowcolor{gray!15}\multicolumn{4}{@{}l}{\textit{Inference}} \\
Executed actions & \multicolumn{3}{c}{1 (OGBench), 8 (LIBERO)} \\
Euler steps & \multicolumn{3}{c}{10} \\
\bottomrule
\end{tabular}
\end{table}

\subsection{SplatWAM-\texorpdfstring{$\mathcal{P}$}{P}}
\label{sec:app_pol_impl}
\begin{figure}[!htb]
\centering
\includegraphics[width=\linewidth]{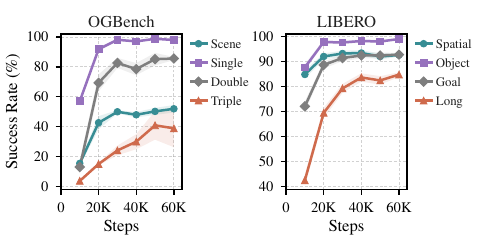}
\caption{
    \textbf{SplatWAM-$\mathcal{P}$ success rate during policy training on OGBench and LIBERO.} Shaded regions show the standard deviation over three training seeds on OGBench and the standard error over evaluation episodes on LIBERO.}
\label{fig:sr_steps}
\iflatexml\end{figure}

For OGBench, we tokenize the goal observation with SplatWAM-$\mathcal{T}$ and process the resulting state in a separate pass through the shared backbone. At each layer, agent tokens attend to the corresponding goal activations as additional keys and values. For LIBERO, we instead use the token outputs of a frozen CLIP~\cite{clip} text encoder, linearly projected as keys and values for the agent tokens. SplatWAM-$\mathcal{P}$ is initialized from a SplatWAM-$\mathcal{D}$ checkpoint after 40K steps and jointly trained with it (Fig.~\ref{fig:app_joint}) using Table~\ref{tab:app_pol_hparams}. Figure~\ref{fig:sr_steps} shows policy success over training, and Table~\ref{tab:per_task} reports per-task success at 60K. SplatWAM-$\mathcal{P}$ exceeds the model-free Gaussian-token policy (Table~\ref{tab:tokbc}) on average, with the largest gains on Scene and Cube-Double.

At each step, SplatWAM-$\mathcal{T}$ tokenizes the current multi-view observation. The policy processes the latest $T$ observed states with null action tokens and samples an action chunk. Observed states use $\tau=1-d_{\min}$ and step size $d_{\min}$ during both training and inference. We execute one action on OGBench or eight on LIBERO before replanning. Longer open-loop chunks reduce success on OGBench, where play data contain temporally correlated action noise~\cite{ogbench}. On LIBERO, chunked execution handles pauses and other non-Markovian behavior in human demonstrations~\cite{diffusionpolicy,act}. The policy conditions only on observed history, not imagined rollouts.

\begin{figure}[t]
\centering
\includegraphics[width=\linewidth]{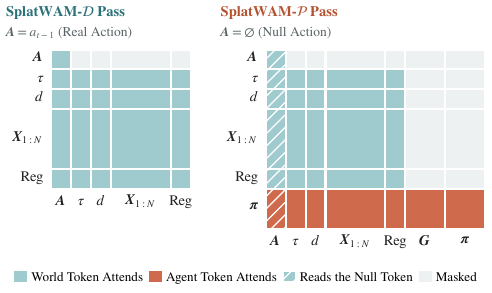}
\caption{\textbf{Within-frame attention masks for the dynamics and policy passes.} The dynamics pass uses the previous action, while the policy pass uses a null action token and adds agent and goal tokens.}
\label{fig:attn}
\end{figure}

Figure~\ref{fig:attn} shows the within-frame attention masks, with queries in rows and keys in columns; cross-frame attention is omitted. In the dynamics pass, the action token $A$ contains $a_{t-1}$ and attends only to itself, while world tokens attend to one another and to $A$. In the policy pass, $A$ is replaced by the learned null token. Agent tokens additionally attend to world tokens, goal tokens, and each other, while world tokens cannot attend to agent or goal tokens.

\begin{table}[!htb]\centering\def\captionof#1{\caption}\else\vspace{8pt}\fi
\scriptsize
\setlength{\tabcolsep}{2.5pt}
\newcommand{\pmv}[1]{{\tiny$\pm$#1}}
\captionof{table}{\textbf{SplatWAM-$\mathcal{P}$ per-task success rate (\%) on \mbox{OGBench} at 60K.} Mean $\pm$ standard deviation over three training seeds. Task indices follow the official OGBench task ordering.}
\label{tab:per_task}
\begin{tabular*}{\columnwidth}{@{\extracolsep{\fill}}lcccccc@{}}
\toprule
Environment & Task 1 & Task 2 & Task 3 & Task 4 & Task 5 & Avg. \\
\midrule
Scene & 100.0\pmv{0.0} & 66.7\pmv{12.7} & 90.7\pmv{3.1} & 2.7\pmv{3.1} & 0.0\pmv{0.0} & 52.0\pmv{2.5} \\
Cube-Single & 98.0\pmv{0.0} & 96.7\pmv{1.2} & 98.7\pmv{1.2} & 97.3\pmv{3.1} & 99.3\pmv{1.2} & 98.0\pmv{1.1} \\
Cube-Double & 96.7\pmv{5.8} & 99.3\pmv{1.2} & 99.3\pmv{1.2} & 35.3\pmv{10.1} & 97.3\pmv{2.3} & 85.6\pmv{1.1} \\
Cube-Triple & 98.7\pmv{1.2} & 40.7\pmv{29.1} & 33.3\pmv{23.2} & 16.7\pmv{14.0} & 5.3\pmv{1.2} & 38.9\pmv{12.6} \\
\bottomrule
\end{tabular*}
\iflatexml\end{table}\else\end{figure}\fi

\begin{table}[!htb]
\centering
\footnotesize
\caption{\textbf{SplatWAM-$\mathcal{P}$ hyperparameters.}}
\label{tab:app_pol_hparams}
\begin{tabular}{@{}lcc@{}}
\toprule
Hyperparameter & OGBench & LIBERO \\
\midrule
\rowcolor{gray!15}\multicolumn{3}{@{}l}{\textit{Architecture}} \\
Goal & goal state $s_g$ & instruction \\
Window length $T$ & \multicolumn{2}{c}{17} \\
Agent tokens & \multicolumn{2}{c}{4} \\
Action head & \multicolumn{2}{c}{flow matching, MLP $2\times512$} \\
\midrule
\rowcolor{gray!15}\multicolumn{3}{@{}l}{\textit{Objective}} \\
Joint objective & \multicolumn{2}{c}{$\mathcal{L}_{\pi}+\mathcal{L}_{\mathrm{SF}}$} \\
Action chunk $k$ & 8 & 16 \\
Goal horizon $\Delta_{\max}$ & 100 & -- \\
\midrule
\rowcolor{gray!15}\multicolumn{3}{@{}l}{\textit{Optimization}} \\
Initialization & \multicolumn{2}{c}{SplatWAM-$\mathcal{D}$ after 40K steps} \\
Optimizer & \multicolumn{2}{c}{AdamW} \\
Learning rate, policy & \multicolumn{2}{c}{$3\times10^{-4}$} \\
Learning rate, world-model layers & \multicolumn{2}{c}{$10^{-4}$} \\
Warmup & \multicolumn{2}{c}{1K steps} \\
LR schedule & \multicolumn{2}{c}{cosine annealing} \\
Batch size & \multicolumn{2}{c}{8} \\
Training steps & \multicolumn{2}{c}{60K} \\
EMA decay & \multicolumn{2}{c}{0.999} \\
\midrule
\rowcolor{gray!15}\multicolumn{3}{@{}l}{\textit{Inference}} \\
Executed actions & 1 & 8 \\
Euler steps & \multicolumn{2}{c}{10} \\
\bottomrule
\end{tabular}
\end{table}

\subsection{Rendering loss during joint training}
\label{sec:app_joint}
Figure~\ref{fig:app_joint} compares continued world-model training with three joint-training variants, all initialized from the same 40K-step SplatWAM-$\mathcal{D}$ checkpoint. With the main-paper objective $\mathcal{L}_\pi+\mathcal{L}_{\mathrm{SF}}$, 16-step rollout PSNR on OGBench Scene drops from 24.02 to 22.00\,dB after 60K policy-training steps. Retaining the rendering loss ($\mathcal{L}_\pi+\mathcal{L}_{\mathrm{dyn}}$) improves PSNR to 22.61\,dB while maintaining policy success (54.5\% vs.\ 52.0\%). Blocking policy gradients ($\mathcal{L}^{\mathrm{sg}}_\pi+\mathcal{L}_{\mathrm{dyn}}$) preserves predictive fidelity (24.36\,dB) but reduces policy success to 46.7\%.

Similar gradient interference has been observed in VLAs~\cite{knowinsulate}, while a gradient-routing study shows that blocking these gradients can hurt control without discrete action-token supervision~\cite{gradrouting}. As in Dreamer~4~\cite{dreamer4}, retaining prediction supervision helps preserve pretrained capabilities. Here, the rendering loss mitigates degradation while allowing policy gradients to update the shared world model.

\subsection{Context views for model-free policies}
\label{sec:app_repcontrol}
\begin{figure}[!t]
\centering
\includegraphics[width=\linewidth]{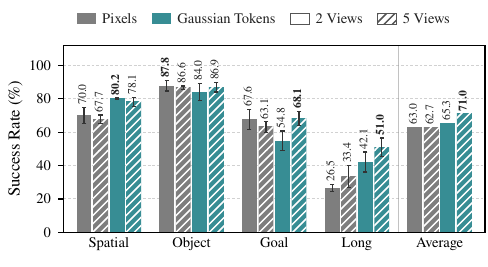}
\caption{\textbf{Effect of context views on model-free policies.} LIBERO success rates for pixel and Gaussian-token policies using 2 or 5 views. Results are averaged over three training seeds.}
\label{fig:app_views}
\end{figure}
Figure~\ref{fig:app_views} extends the two-view comparison of Table~\ref{tab:tokbc} to five context views. Additional views improve the Gaussian-token policy's average LIBERO success from 65.3\% to 71.0\%, while the pixel policy remains nearly unchanged (63.0\% to 62.7\%). The largest gains occur on Goal (+13.3 points) and Long (+8.9 points), where small or partially occluded objects require more precise reconstruction. With five views, Gaussian tokens match or outperform pixels on every suite. These results suggest that sparse-view reconstruction limits Gaussian-token control on cluttered tasks.

\subsection{Gaussian tokens vs.\ VGGT features}
\label{sec:app_vggt}
\begin{figure}[!t]
\centering
\includegraphics[width=\linewidth]{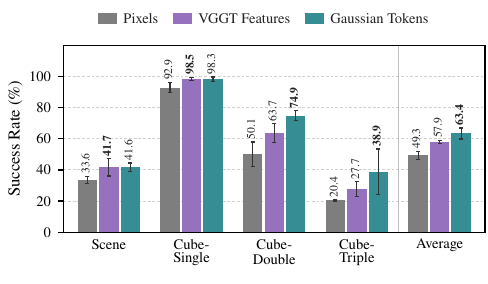}
\caption{\textbf{Gaussian tokens vs.\ VGGT features.} Model-free success rate on OGBench using two context views. VGGT features are extracted from SplatWAM-$\mathcal{T}$'s fine-tuned backbone. Results are averaged over three training seeds.}
\label{fig:app_vggt}
\end{figure}
\begin{table}[!htb]
\centering
\scriptsize
\setlength{\tabcolsep}{2.5pt}
\newcommand{\pmv}[1]{{\tiny$\pm$#1}}
\caption{\textbf{Per-task success rate (\%) on OGBench Cube-Double.} Mean $\pm$ standard deviation over three training seeds. PnP denotes pick-and-place.}
\label{tab:app_vggt}
\begin{tabular*}{\columnwidth}{@{\extracolsep{\fill}}lcccccc@{}}
\toprule
& \shortstack{Single\\PnP} & \shortstack{Double\\PnP 1} & \shortstack{Double\\PnP 2} & Swap & Stack & Avg. \\
\midrule
Pixels & 98.0\pmv{3.5} & 58.0\pmv{9.2} & 38.7\pmv{11.0} & 0.0\pmv{0.0} & 56.0\pmv{40.8} & 50.1\pmv{7.8} \\
VGGT Features & 98.7\pmv{1.2} & 84.7\pmv{9.9} & 51.3\pmv{15.0} & 0.0\pmv{0.0} & \textbf{84.0}\pmv{26.0} & 63.7\pmv{6.1} \\
Gaussian Tokens & \textbf{99.3}\pmv{1.2} & \textbf{100.0}\pmv{0.0} & \textbf{100.0}\pmv{0.0} & \textbf{9.3}\pmv{14.5} & 66.0\pmv{15.9} & \textbf{74.9}\pmv{3.2} \\
\bottomrule
\end{tabular*}
\end{table}
To distinguish the benefit of Gaussian representations from visual pretraining, we replace the Gaussian tokens of the model-free policy in Table~\ref{tab:tokbc} with features from SplatWAM-$\mathcal{T}$'s fine-tuned VGGT backbone (Fig.~\ref{fig:app_vggt}). We retain all $2\times256$ patch tokens attended to by the Gaussian queries and reduce them to 48 dimensions using PCA, preserving 98\% of the variance. Both policies use the same Set Transformer and training protocol (Table~\ref{tab:app_mf_hparams}). A linear probe recovers patch positions with $R^2\geq0.99$, showing that spatial position remains linearly recoverable from VGGT features.

VGGT features outperform pixels, consistent with prior findings on pretrained visual representations for control~\cite{hansenscratch,vc1,spa}. Gaussian tokens improve further on average (Fig.~\ref{fig:app_vggt}), especially on double pick-and-place (Table~\ref{tab:app_vggt}). We hypothesize that their explicit 3D positions in a shared coordinate frame make it easier to compare current and goal configurations, as explored with spatial feature points~\cite{dsae} and Gaussian-splat states~\cite{gsrl}. VGGT features perform better on stacking, but the difference is within seed variation.

\end{document}